\documentclass[runningheads]{llncs}
\usepackage[pagebackref=true,breaklinks=true,letterpaper=true,colorlinks,bookmarks=false]{hyperref}

\usepackage{graphicx}
\usepackage{comment}
\usepackage{amsmath,amssymb} 
\usepackage{color}
\usepackage{mathtools}
\usepackage{multirow}
\usepackage{wrapfig}
\usepackage{subcaption}
\usepackage{forloop}
\usepackage{forarray}
\usepackage{floatrow}

\newfloatcommand{capbtabbox}{table}[][\FBwidth]
\usepackage[strings]{underscore}
\usepackage[width=122mm,left=12mm,paperwidth=146mm,height=193mm,top=12mm,paperheight=217mm]{geometry}

\graphicspath{{figures/}}
\newcommand{\orcid}[1]{\href{https://orcid.org/#1}{\includegraphics[width=10pt]{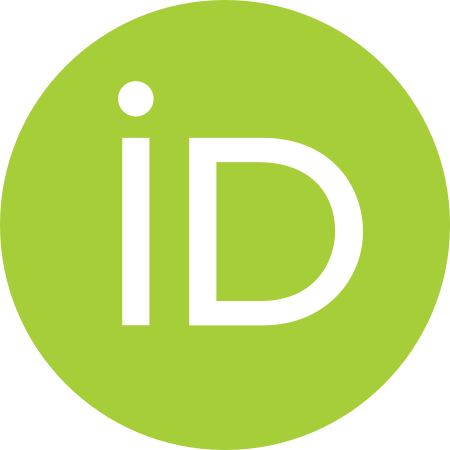}}}
\begin{document}
\pagestyle{headings}
\mainmatter
\def\ECCVSubNumber{6753}  

\title{Synthesizing Coupled 3D Face Modalities by Trunk-Branch Generative Adversarial Networks} 

\titlerunning{Synthesizing Coupled 3D Face Modalities by TBGAN}
%
\author{Baris Gecer\inst{1,2}\thanks{Corresponding author}\orcid{0000-0002-5684-2843} \and
Alexandros Lattas\inst{1,2}\orcid{0000-0002-9964-6105} \and
Stylianos Ploumpis\inst{1,2}\orcid{0000-0002-4836-1513} \and\\
Jiankang Deng\inst{1,2}\orcid{0000-0002-3709-6216} \and
Athanasios Papaioannou\inst{1,2}\orcid{0000-0002-5665-3305} \and
Stylianos Moschoglou\inst{1,2}\orcid{0000-0001-7421-1335} \and
Stefanos Zafeiriou\inst{1,2}\orcid{0000-0002-5222-1740}}

\authorrunning{B. Gecer et al.}
%
\institute{Imperial College, London, UK \\ \url{https://ibug.doc.ic.ac.uk/}\\
\email{\{b.gecer, a.lattas, s.ploumpis, j.deng16, a.papaioannou11, stylianos.moschoglou15, s.zafeiriou\}@imperial.ac.uk}\\
\and FaceSoft.io, London, UK}
\maketitle

\begin{abstract}
Generating realistic 3D faces is of high importance for computer graphics and computer vision applications. Generally, research on 3D face generation revolves around linear statistical models of the facial surface. Nevertheless, these models cannot represent faithfully either the facial texture or the normals of the face, which are very crucial for photo-realistic face synthesis. Recently, it was demonstrated that Generative Adversarial Networks (GANs) can be used for generating high-quality textures of faces. Nevertheless, the generation process either omits the geometry and normals, or independent processes are used to produce 3D shape information. In this paper, we present the first methodology that generates high-quality texture, shape, and normals jointly, which can be used for photo-realistic synthesis. To do so, we propose a novel GAN that can generate data from different modalities while exploiting their correlations. Furthermore, we demonstrate how we can condition the generation on the expression and create faces with various facial expressions. The qualitative results shown in this paper are compressed due to size limitations, full-resolution results and the accompanying video can be found in the supplementary documents. The code and models are available at the project page: \url{https://github.com/barisgecer/TBGAN}.
\end{abstract}

\section{Introduction}

\begin{figure*}[t]\centering
\includegraphics[width=1.0\textwidth]{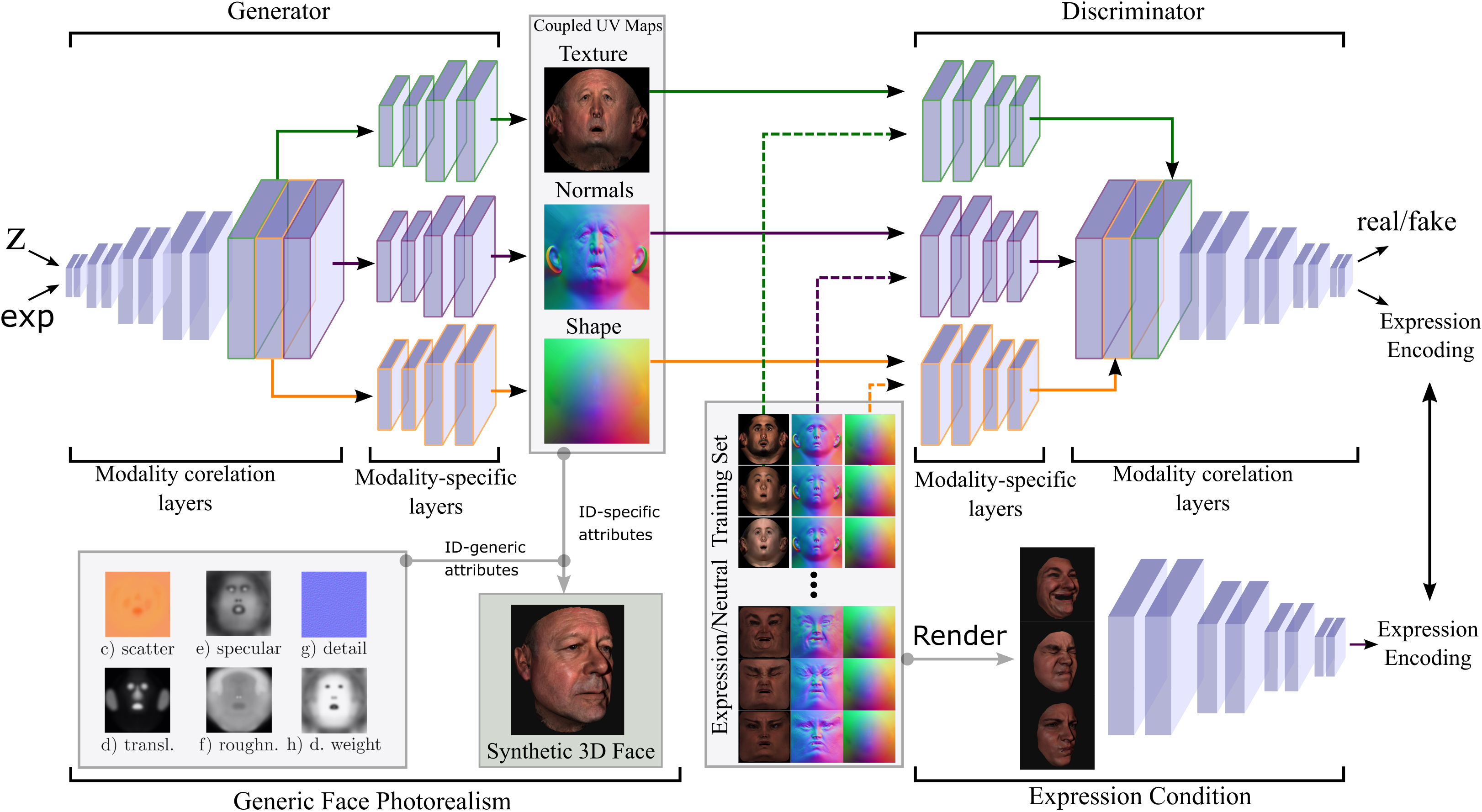}
\caption{We propose a novel GAN that can synthesize high-quality texture, shape, and normals jointly for realistic and coherent 3D faces of novel identities. The separation of branch networks allows the specialization of the characteristic of each one of the modalities while the trunk network maintains the local correspondences among them. Moreover, we demonstrate how we can condition the generation on the expression and create faces with various facial expressions. We annotate the training dataset automatically by an expression recognition network to couple those expression encodings to the texture, shape, and normals UV maps.}
\label{fig:teaser}
\end{figure*}

Generating 3D faces with high-quality texture, shape, and normals is of paramount importance in computer graphics, movie post-production, computer games, etc. Other applications of such approaches include generating synthetic training data for face recognition~\cite{gecer2018facegan} and modeling the face manifold for 3D face reconstruction~\cite{gecer2019ganfit}.  Currently, 3D face generation in computer games and movies is performed by expensive capturing systems or by professional technical artists. The current state-of-the-art methods generate faces, which can be suitable for applications such as caricature avatar creation in mobile devices~\cite{hu2017avatar} but do not generate high-quality shape and normals that can be used for photo-realistic face synthesis. In this paper, we propose the first methodology for high-quality face generation that can be used for photo-realistic face synthesis (i.e., joint generation of texture, shape, and normals) by capitalizing on the recent developments on Generative Adversarial Networks (GANs). 

The early face models, such as \cite{blanz1999morphable}, represent 3D face by disentangled PCA models of geometry, expression~\cite{cao2013facewarehouse}, and colored texture, called 3D morphable models (3DMM). 3DMMs and its variants were the most popular method for modeling shape and texture separately. However, the linear nature of PCA is often unable to capture high-frequency signals properly, thus the quality of generation and reconstruction by PCA is sub-optimal.

GANs is a recently introduced family of techniques that train samplers of high-dimensional distributions~\cite{goodfellow2014generative}. It has been demonstrated that when a GAN is trained on facial images, it can generate images that have realistic characteristics. In particular, the recently introduced GANs~\cite{karras2017progressive,karras2018style,brock2018large} can generate photo-realistic high-resolution faces. Nevertheless, because they are trained on partially-aligned 2D images, they cannot properly model the manifold of faces and thus (a) inevitably create many unrealistic instances and (b) it is not clear how they can be used to generate photo-realistic 3D faces. 

Recently, GANs have been applied for generating facial texture for various applications. In particular, \cite{sela2017unrestricted} and \cite{gecer2018facegan} utilize style transfer GANs to generate photorealistic images of 3DMM-sampled novel identities. \cite{slossberg2018high} directly generates high-quality 3D facial textures by GANs and \cite{gecer2019ganfit} replaces 3D Morphable Models (3DMMs) with GAN models for 3D texture reconstruction while the shape is still maintained by statistical models. \cite{lattas2020avatarme} propose to generate  4K diffuse and specular albedo and normals from a texture map by an image-to-image GAN. On the other hand, \cite{moschoglou20193dfacegan} model 3D shape by GANs in a parametric UV map and \cite{ranjan2018generating} utilize mesh convolutions with variational autoencoders to model shape in its original structure. Although one can model 3D faces with such shape and texture GAN approaches, these studies omit the correlation between shape, normals, and texture which is very important for photorealism in identity space. The significance of such correlation is most visible with inconsistent facial attributes such as age, gender, and ethnicity (i.e. old-aged texture on a baby-face geometry).

In order to address these gaps, we propose a novel multi-branch GAN architecture that preserves the correlation between different 3D modalities (such as texture, shape, normals, and expression). After converting all modalities into UV space and concatenate over channels, we train a GAN that generates all modalities in a meaningful local and global correspondence. In order to prevent incompatibility issues due to the intensity distribution of different modalities, we propose a trunk-branch architecture that can synthesize photorealistic 3D faces with coupled texture and geometry. Further, we condition this GAN by expression labels to generate faces in any desired expression.

From a computer graphics point of view, a photorealistic face rendering requires a number of elements to be tailored, i.e. shape, normals and albedo maps, some of which should or can be specific to a particular identity. However, the cost of hand-crafting novel identities limits their usage on large-scale applications. The proposed approach tackles this down with reasonable photorealism with a massively generalized identity space. Although the results in this paper are limited to aforementioned modalities by the dataset at hand, the proposed method allows adding more identity-specific modalities (i.e. cavity, gloss, scatter) once such a dataset becomes available.

The contributions of this paper can be summarized as follows:
\begin{itemize}
\item We propose to model and synthesize coherent 3D faces by jointly training a novel Trunk-branch based GAN (TBGAN) architecture for shape, texture, and normals modalities. TBGAN is designed to maintain correlation while tolerating domain-specific differences of these three modalities and can be easily extended to other modalities and domains.
\item In the domain of identity-generic face modeling, we believe this is the first study that utilizes normals as an additional source of information.
\item We propose the first methodology for face generation that correlates expression and identity geometries (i.e. modeling personalized expression) and also the first attempt to model expression in texture and normals space.
\end{itemize}

\section{Related Work}
\subsection{3D face modeling}
There is an underlying assumption that human faces lie on a manifold with respect to the appearance and geometry. As a result, one can model the geometry and appearance of the human face analytically based upon the identity and expression space of all individuals. Two of the first attempts in the history of face modeling were \cite{akimoto1993automatic}, which proposes part-based 3D face reconstruction from frontal and profile images, and \cite{platt1981animating}, which represents expression action units by a set of muscle fibers. 

Twenty years ago methods that generated 3D faces revolved around parametric generative models that are driven by a small number of anthropometric statistics (e.g., sparse face measurements in a population) which act as constraints~\cite{decarlo1998anthropometric}. The seminal work of 3D morphable models (3DMMs)~\cite{blanz1999morphable} demonstrated for the first time that is possible to learn a linear statistical model from a population of 3D faces~\cite{patel20093d,brunton2014review}. 3DMMs are often constructed by using a Principal Component Analysis (PCA) based on a dataset of registered 3D scans of hundreds~\cite{paysan20093d} or thousands~\cite{booth2018large} subjects. Similarly, facial expressions are also modeled by applying PCA~\cite{yang2011expression,li2017learning,breidt2011robust,amberg2008expression}, or are manually defined using linear blendshapes~\cite{li2010example,thies2015real,bouaziz2013online}. 3DMMs, despite their advantages, are bounded by the capacity of linear space that under-represents the high-frequency information and often result in overly-smoothed geometry and texture models. \cite{chen2019photo} and \cite{tran2018extreme} attempt to address this issue by using local displacement maps. Furthermore, the 3DMM line of research assumes that texture and shape are uncorrelated, hence they can only be produced by separate models (i.e., separate PCA models for texture and shape). Early attempts in correlated shape and texture have been made in Active Appearance Models (AAMs) by computing joint PCA models of sparse shape and texture~\cite{cootes2001active}. Nevertheless, due to the inherent limitations of PCA to model high-frequency texture, it is rarely used to correlate shape and texture for 3D face generation. 

Recent progress in generative models~\cite{kingma2013auto,goodfellow2014generative} is being utilized in 3D face modeling to tackle this issue. \cite{moschoglou20193dfacegan} trained a GAN that models face geometry based on UV representations for neutral faces, and likewise, \cite{ranjan2018generating} modeled identity and expression geometry by variational autoencoders with mesh convolutions. \cite{gecer2019ganfit} proposed a GAN-based texture modeling for 3D face reconstruction while modeling geometry by PCA and \cite{slossberg2018high} trained a GAN to synthesize facial textures. To the best of our knowledge, these methodologies totally omit the correlation between geometry and texture and moreover, they ignore identity-specific expression modeling by decoupling them into separate models. In order to address this issue, we propose a trunk-branch GAN that is trained jointly for texture, shape, normals, and expression in order to leverage non-linear generative networks for capturing the correlation between these modalities.

\subsection{Photorealistic face synthesis}

Although most of the aforementioned 3D face models can synthesize 2D face images, there are also some dedicated 2D face generation studies. \cite{mohammed2009visio} combines non-parametric local and parametric global models to generate various set of face images. Recent family of GAN approaches~\cite{radford2015unsupervised,karras2017progressive,karras2018style,brock2018large} offers the state-of-the-art high quality random face generation without constraints.

Some other GAN-based studies allow to condition synthetic faces by rendered 3DMM images~\cite{gecer2018facegan}, by landmarks~\cite{bazrafkan2018face} or by another face image~\cite{bao2018towards} (i.e. by disentangling identity and certain facial attributes). Similarly, facial expression is also conditionally synthesized by an audio input~\cite{jamaludin2019you}, by action unit codes~\cite{pumarola2018ganimation}, by predefined 3D geometry~\cite{zhang2005geometry} or by expression of an another face image~\cite{li2012data}.

In this work, we jointly synthesize the aforementioned modalities for coherent photorealistic face synthesis by leveraging high-frequency generation by GANs. Unlike many of its 2D and 3D alternatives, the resulting generator models provide absolute control over disentangled identity, pose, expression and illumination spaces. Unlike many other GAN works that are struggling due to misalignments among the training data, our entire latent space correspond to realistic 3D faces as the data representation is naturally aligned on UV space.
		
\subsection{Boosting face recognition by synthetic training data}
There have been also some works to synthesize face images to be used as synthetic training data for face recognition methods either by directly using GAN-generated images~\cite{trigueros2018generating} or by controlling pose-space with a conditional-GAN~\cite{tran2018representation,hu2018pose,shen2018faceid}. \cite{masi2016we} propose many augmentation techniques, such as rotation, expression, and shape, based on 3DMMs. Other GAN-based approaches that capitalize 3D facial priors include \cite{zhao2017dual}, which rotates faces by fitting 3DMM and preserves photorealism by translation GANs and \cite{yin2017towards}, which frontalize face images by a GAN and 3DMM regression network. \cite{deng2018uv} complete missing parts of UV texture representations of 2D images after 3DMM fitting by a translation GAN. \cite{gecer2018facegan} first synthesizes face images of novel identities by sampling from 3DMM and then removes the photorealistic domain gap by an image-to-image translation GAN.

All of these studies show the significance of photorealistic and identity-generic face synthesization for the next generation of facial recognition algorithms. Although this study focuses more on the graphical aspect of face synthesization, we show that synthetic images can also improve face recognition performance.


\section{Approach}

\subsection{UV Maps for Shape, Texture and Normals}
\label{sec:UV-section}

\begin{figure}[t]
\begin{center}

\includegraphics[trim={0 12cm 0 0},clip,width=0.35\linewidth]{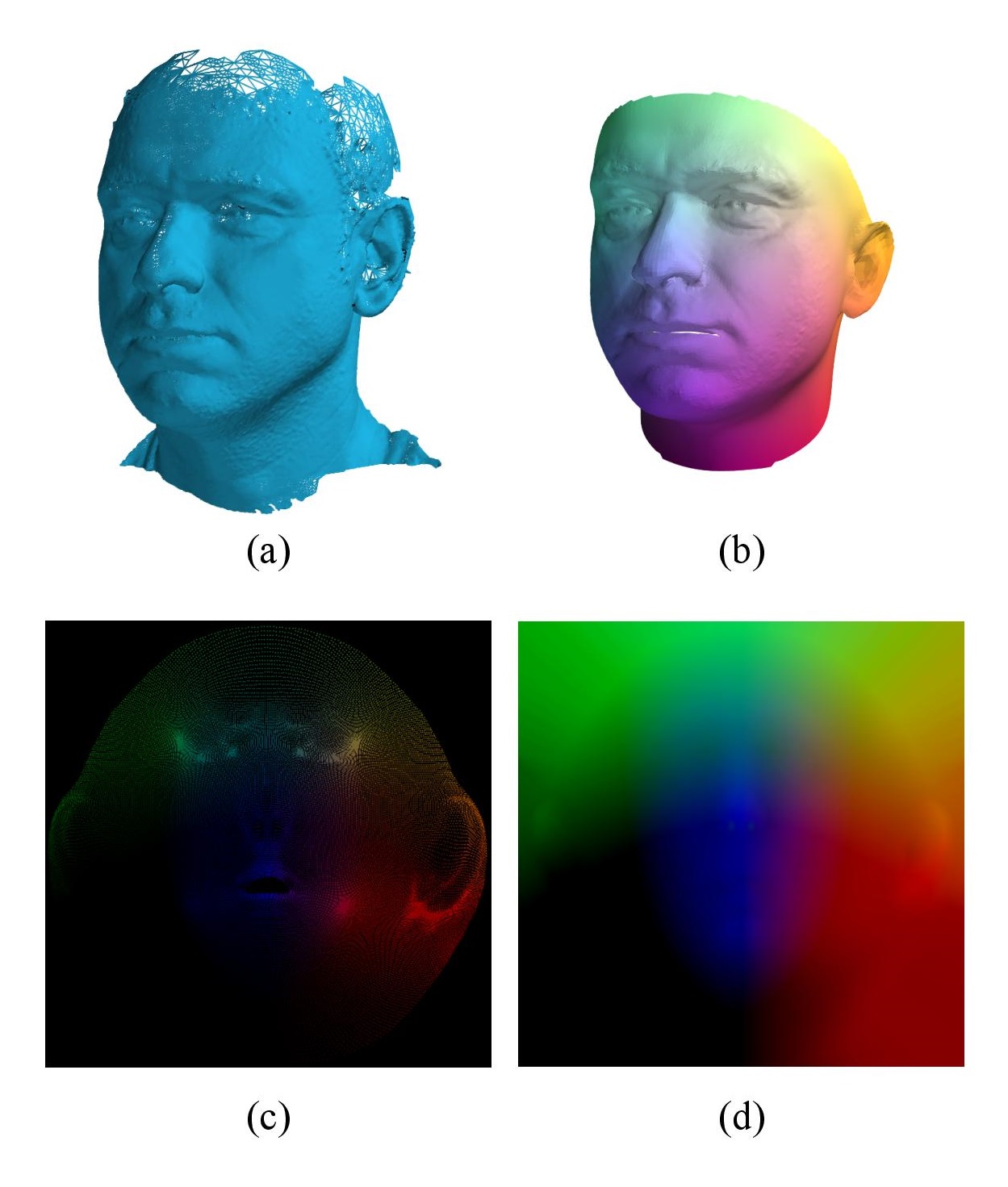}
\includegraphics[trim={0 0 0 12cm},clip,width=0.35\linewidth]{main/inter}
\vspace{-0.5cm}
\caption{UV extraction process. In (a) we present a raw mesh, in (b)  the registered mesh using the Large Scale Face Model (LSFM) template~\cite{booth20163d}, in (c) the unwrapped 3D mesh in the 2D UV space, and (d) the interpolated 2D UV map. Interpolation is carried out using the barycentric coordinates of each pixel in the registered 3D mesh.}
\label{fig:uv-extraction}
\end{center}
\end{figure}

In order to feed the shape, the texture, and the normals of the facial meshes into a deep network we need to reparameterize them into an image-like tensor format to apply 2D-convolutions~\footnote{Another line of research is mesh convolutional networks~\cite{cheng2019meshgan,litany2018deformable,ranjan2018generating} which cannot preserve high-frequency details of the texture and normals at the current state-of-the-art.}. We begin by describing all the raw 3D facial scans with the same topology and number of vertices (dense correspondence). This is achieved by morphing non-rigidly a template mesh to each one of the raw scans. We employ a standard non-rigid iterative closest point algorithm as described in \cite{amberg2007optimal,de2010optimal} and we deform our chosen template so that it captures correctly the facial surface of the raw scans. As a template mesh, we choose the mean face of the LSFM model proposed in \cite{booth20163d}, which consists approximately of $54K$ vertices that are sufficient enough to depict non-linear, high facial details.

After reparameterizing all the meshes into the LSFM~\cite{booth20163d} topology, we cylindrically unwrap the mean face of the LSFM~\cite{booth20163d} to create a UV representation for that specific mesh topology. In the literature, a UV map is commonly utilized for storing only the RGB texture values. Apart from storing the texture values of the 3D meshes, we utilize the UV space to store the 3D coordinates of each vertex $(x,y,z)$ and the normal orientation $(n_x,n_y,n_z)$. Before storing the 3D coordinates into the UV space, all meshes are aligned in the 3D spaces by performing General Procrustes Analysis (GPA)~\cite{gower1975generalized} and are normalized to be in the scale of $[1, -1]$. Moreover, we store each 3D coordinate and normals in the UV space given the respective UV pixel coordinate. Finally, we perform a barycentric interpolation based on the barycentric coordinates of each pixel on the registered mesh to fill out the missing areas in order to produce a dense illustration of the UV map. In Fig. \ref{fig:uv-extraction}, we illustrate a raw 3D scan, the registered 3D scan on the LSFM~\cite{booth20163d} template, the sparse UV map of 3D coordinates and finally the interpolated one.


\subsection{Trunk-Branch GAN to Generate Coupled Texture, Shape and Normals}


In order to train a model that handles multiple modalities, we propose a novel trunk-branch GAN architecture to generate entangled modalities of the 3D face such as texture, shape, and normals as UV maps. For this task, we exploit the MeIn3D dataset~\cite{booth20163d} which consists of approximately 10,000 neutral 3D facial scans with wide diversity in age, gender, and ethnicity.

Given a generator network $\mathcal{G}^L$ with a total of $L$ convolutional upsampling layers and gaussian noise $\mathbf{z}\sim\mathcal{N}\left(\mathbf{0},\mathbf{I}\right)$ as input, the activation at the end of layer $d$ (i.e., $\mathcal{G}^d(\mathbf{z})$) is split into three branch networks $\mathcal{G}_T^{L-d}$, $\mathcal{G}_N^{L-d}$, $\mathcal{G}_S^{L-d}$ each of which consists of $L-d$ upsampling convolutional layers that generate texture, normals and shape UV maps respectively. The discriminator $\mathcal{D}^L$ starts with the branch networks $\mathcal{D}_T^{L-d}$, $\mathcal{D}_N^{L-d}$, $\mathcal{D}_S^{L-d}$ whose activations are concatenated before fed into trunk network $\mathcal{D}^d$. The output of $\mathcal{D}^L$ is regression of real/fake score. 

Although the proposed approach is compatible with most of the GAN architectures and loss functions, in our experiments, we base TBGAN on progressive growing GAN architecture~\cite{karras2017progressive} train it  by WGAN-GP loss~\cite{gulrajani2017improved} as following:
\begin{align}
\mathcal{L}_{\mathcal{G}^L} &= \mathbb{E}_{\mathbf{z}\sim\mathcal{N}\left(\mathbf{0},\mathbf{I}\right)} \left[ - \mathcal{D}^L\left( \mathcal{G}^L(\mathbf{z}) \right)\right] \label{eq:tbgan-generator-loss} \\ 
\mathcal{L}_{\mathcal{D}^L} &= \mathbb{E}_{x \sim p_{\text{data}},~\mathbf{z}\sim\mathcal{N} \left(\mathbf{0},\mathbf{I}\right)} \left[\mathcal{D}^L\left(\mathcal{G}^L(\mathbf{z})\right) -\mathcal{D}^L(x) + \lambda * G P(x, \mathcal{G}^L(\mathbf{z})) \right] \label{eq:tbgan-discriminator-loss}
\end{align}
where gradient penalty calculated by $G P(x, \hat{x}) = \left( \| \nabla \mathcal{D}^L\left(\alpha \hat{x} + (1-\alpha) x\right) \|_{2}-1 \right)^{2}$ and $\alpha$ denotes uniform random numbers between 0 and 1. $\lambda$ is a balancing factor which is typically $\lambda= 10$. An overview of this trunk-branch architecture is illustrated in Fig.~\ref{fig:teaser}

\subsection{Expression Augmentation by Conditional GAN}


Further, we modify our GAN in order to generate 3D faces with expression by conditioning it with expression annotations ($\mathbf{p_e}$). Similar to the MeIn3D dataset, we have captured approximately $35,000$ facial scans of around $5,000$ distinct identities during a special exhibition in the Science Museum, London. All subjects were recorded in various guided expressions with a 3dMD face capturing apparatus. All of the subjects were asked to provide meta-data regarding their age, gender, and ethnicity. The database consists of $46\%$ male, $54\%$ female, $85\%$ White, $7\%$ Asian, $4\%$ Mixed Heritage, $3\%$ Black, and $1\%$ other. 

In order to avoid the cost and potential inconsistency of manual annotation, we render those scans and automatically annotate them by an expression recognition network. The resulting expression encodings ($(*,\mathbf{p_e})  \sim p_{\text { data }}$) are used as label vector during the training of our trunk-branch conditional GAN. This training scheme is illustrated in Fig.~\ref{fig:teaser}. $\mathbf{p_e}$ is basically a vector of 7 for universal expressions (neutral, happy, angry etc.), randomly drawn from our dataset. During the training, Eq.~\ref{eq:tbgan-generator-loss} and \ref{eq:tbgan-discriminator-loss} are updated to condition expression encodings by AC-GAN~\cite{odena2017conditional} as following:
\begin{align}
\mathcal{L}_{\mathcal{G}^L} &\mathrel{+}= \mathbb{E}_{(*,\mathbf{p_e})  \sim p_{\text{data}},~\mathbf{z}\sim\mathcal{N}\left(\mathbf{0},\mathbf{I}\right)} \left[ \sum_e  \mathbf{p_e} \log(\mathcal{D}_e^L(\mathcal{G}^L(\mathbf{z},\mathbf{p_e})))   \right]\\  
\mathcal{L}_{\mathcal{D}^L} &\mathrel{+}=  \mathbb{E}_{(x,\mathbf{p_e})  \sim p_{\text{data}},~\mathbf{z}\sim\mathcal{N}\left(\mathbf{0},\mathbf{I}\right)} 
\!\!\left[ \sum_e \mathbf{p_e} \log( \mathcal{D}_e^L(x) ) + \mathbf{p_e} \log( \mathcal{D}_e^L(\mathcal{G}^L(\mathbf{z},\mathbf{p_e})\!)\!)  \! \right]
\end{align}

\noindent which performs softmax cross entropy between expression prediction of the discriminator ($\mathcal{D}_e^L(x)$) and the random expression vector input ($\mathbf{p_e}$) for real ($x$) and generated samples ($\mathcal{G}^L(\mathbf{z},\mathbf{p_e})$).

Unlike previous expression models that omit the effect of the expression on textures, the resulting generator is capable of generating coupled texture, shape, and normals map of a face with controlled expression. Similarly, our generator respects the identity-expression correlation thanks to correlated supervision provided by the training data. This is in contrast to the traditional statistical expression models which decouples expression and identity models into two separate entities.

\subsection{Photorealistic Rendering with Generated UV maps}
\label{sec:generic}
For the renderings to appear photorealistic, we use the generated identity-specific mesh, texture, and normals, in combination with the generic reflectance properties, and employ a commercial rendering application: \textit{Marmoset Toolbag}~\cite{Marmoset19:Toolbag}.

\begin{figure*}[t]
\centering
    \begin{subfigure}{0.22\linewidth}
        \includegraphics[width=\linewidth]{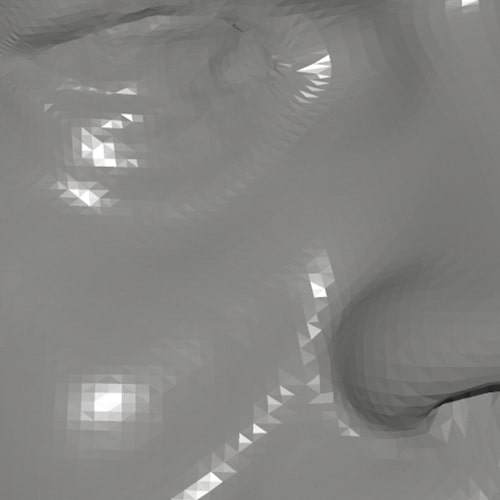}
        \label{fig:rendering-process-results-a}
        \caption{shape}
    \end{subfigure}
    \begin{subfigure}{0.22\linewidth}
        \includegraphics[width=\linewidth]{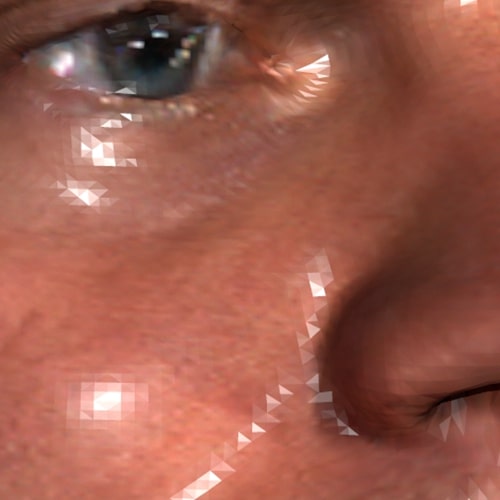}
        \label{fig:rendering-process-results-b}
        \caption{Shp+tex}
    \end{subfigure}
    \begin{subfigure}{0.22\linewidth}
        \includegraphics[width=\linewidth]{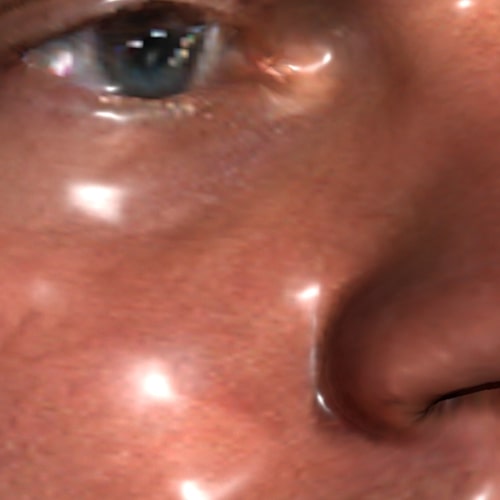}
        \label{fig:rendering-process-results-c}
        \caption{Shp+tex+nor}
    \end{subfigure}
    \begin{subfigure}{0.22\linewidth}
        \includegraphics[width=\linewidth]{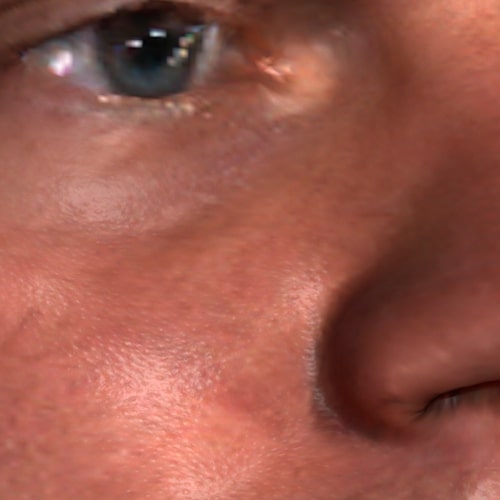}
        \label{fig:rendering-process-results-d}
        \caption{Final}
    \end{subfigure}
   
    \caption{
    Zoom-in on rendering results with
    (a) only the shape,
    (b) adding the albedo texture,
    (c) adding the generated normals,
    and (d) using identity-generic detail normal, specular albedo, roughness, scatter and translucency maps.
    }
    \label{fig:rendering-process-results}
\end{figure*}

In order to extract the 3D representation from the UV domain we employ the inverse procedure explained in section \ref{sec:UV-section} based on the UV pixel coordinates of each vertex of the 3D mesh. Fig.~\ref{fig:rendering-process-results} shows the rendering results, under a single light source, when using the generated geometry (Fig.~\ref{fig:rendering-process-results}(a)) and the generated texture (Fig.~\ref{fig:rendering-process-results}(b)). Here the specular reflection is calculated on the per-face normals of the mesh and exhibits steep changes between on the face's edges. By interpolating the generated normals on each face (Fig.~\ref{fig:rendering-process-results}(c)), we are able to smooth the specular highlights and correct any high-frequency noise on the geometry of the mesh. However, these results do not correctly model the human skin and resemble a metallic surface. 
In reality, the human skin is rough and as a body tissue,
it both reflects and absorbs light, thus exhibiting specular reflection, diffuse reflection, and subsurface scattering.

Although we can add such modalities as additional branches with the availability of such data, we find that rendering can be still improved by adding some identity-generic maps.
Using our training data,
we create maps that define certain reflectance properties per-pixel,
which will match the features of the average generated identity,
as shown in bottom-left of Fig.~\ref{fig:teaser}.
\textit{Scattering} (c) defines the intensity of subsurface scattering of the skin.
\textit{Translucency} (d) defines the amount of light, that travels inside the skin and gets emitted in different directions.
\textit{Specular albedo} (e) gives the intensity of the specular highlights, which differ between hair-covered areas, the eyes, and the teeth.
\textit{Roughness} (f) describes the scattering of specular highlights
and controls the glossiness of the skin.
A \textit{detail normal map} (g) is also tilled and added on the generated normal maps, to mimic the skin pores and a
\textit{detail weight map} (h) controls the appearance of the detail normals, so that they do not appear on the eyes, lips, and hair. The final result (Fig.~\ref{fig:rendering-process-results}(d)) 
properly models the skin surface and reflection,
by adding plausible high-frequency specularity and subsurface scattering,
both weighted by the area of the face where they appear.


\section{Results}
In this section, we give qualitative and quantitative results of our method for generating 3D faces with novel identities and various expressions. In our experiments, there are total $L=8$ up- and down-sampling layers where $d=6$ of them in the trunk and 2 layers in each branch. These choices are empirically validated to ensure sufficient correlation among modalities without incompatibility artifacts. Running time is a few milliseconds to generate UV images from a latent code on a high-end GPU. Transforming from UV image to mesh is just sampling with UV coordinates and can be considered free of cost. Renderings in this paper take a few seconds due to high resolution but this cost depends on the application. The memory needed for the generator network is 1.25GB compared to the 6GB PCA model of the same resolution and $\%95$ of the total variance. 

In the following sections, we first visualize generated UV maps and their contributions to the final renderings on several generated faces. Next, we show the generalization ability of the identity and expression generators on some facial characteristics. We also demonstrate its well-generalization latent space by interpolating between different identities. Additionally, we perform full-head completion to the interpolated faces. Finally, we perform face recognition experiments by using the generated face images as additional training data.

\subsection{Qualitative Results}

\subsubsection{Combining coupled modalities:} Fig.~\ref{fig:uvmaps} presents the generated shape, normals, and texture maps by the proposed GAN and their additive contributions to the final renderings. As can be seen from local and global correspondences, the generated UV maps are highly correlated and coherent. Attributes like age, gender, race, etc. can be easily grasped from all of the UV maps and rendered images. Please also note that some of the minor artifacts of the generated geometry in Fig.~\ref{fig:uvmaps}(d) are compensated by the normals in Fig.~\ref{fig:uvmaps}(e). 

\def \var {0.157}
\begin{figure*}[t!]
\ForEach
{,}
{\includegraphics[width=\var\textwidth]{qualitative/uv_maps/000\thislevelitem_shp}
\includegraphics[width=\var\textwidth]{qualitative/uv_maps/000\thislevelitem_nor}
\includegraphics[width=\var\textwidth]{qualitative/uv_maps/000\thislevelitem}
\includegraphics[width=\var\textwidth]{qualitative/327_shp_0_30_60/000\thislevelitem_30}
\includegraphics[width=\var\textwidth]{qualitative/327_shp_nor_0_30_60/000\thislevelitem_30}
\includegraphics[width=\var\textwidth]{qualitative/327_shp_tex_nor_0_30_60/000\thislevelitem_30}
}{008,013,028}
\def \i {041}
\centering 
\begin{subfigure}{\var\textwidth}
	\includegraphics[width=1.0\textwidth]{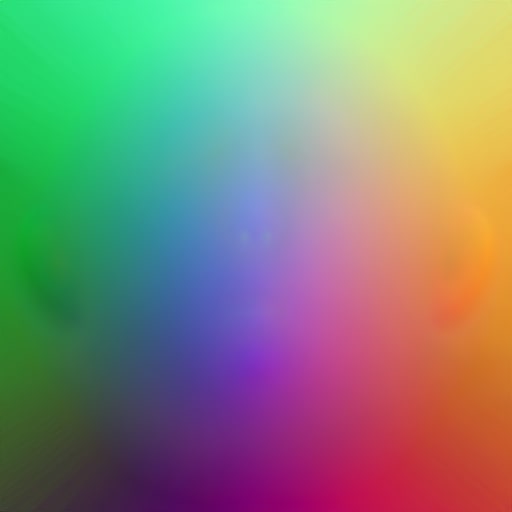}
\caption{Shape}\end{subfigure}
\begin{subfigure}{\var\textwidth}
	\includegraphics[width=1.0\textwidth]{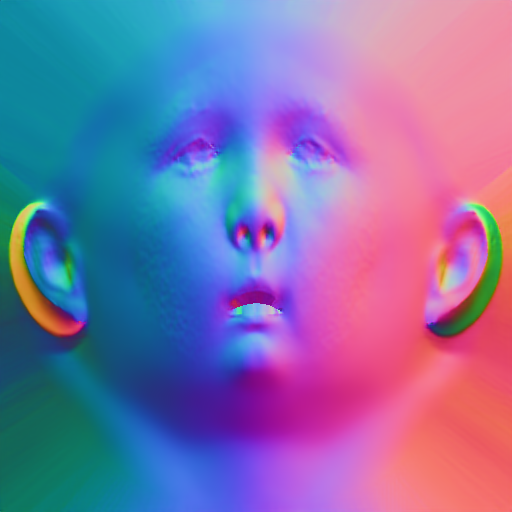}
\caption{Normals}\end{subfigure}
\begin{subfigure}{\var\textwidth}
	\includegraphics[width=1.0\textwidth]{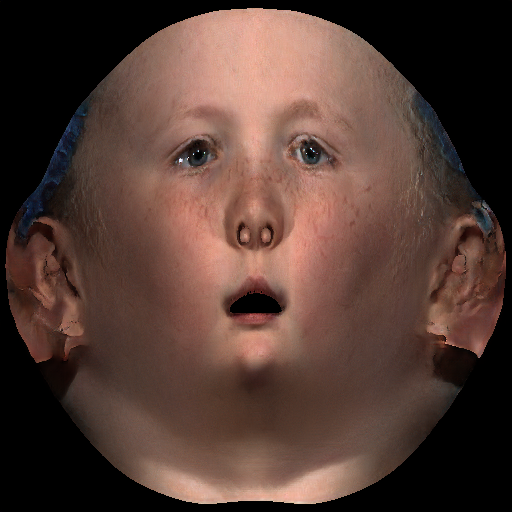}
\caption{Texture}\end{subfigure}
\begin{subfigure}{\var\textwidth}
	\includegraphics[width=1.0\textwidth]{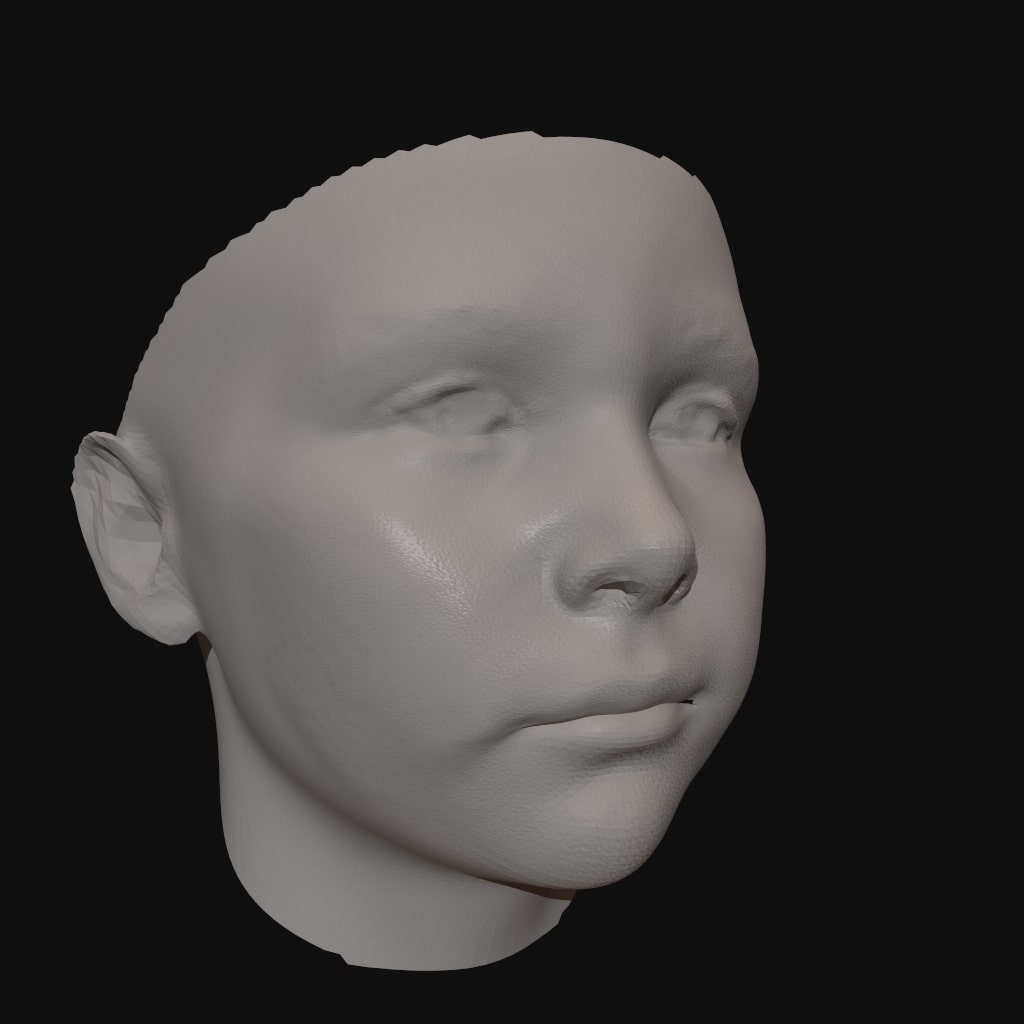}
\caption{Shape}\end{subfigure}
\begin{subfigure}{\var\textwidth}
	\includegraphics[width=1.0\textwidth]{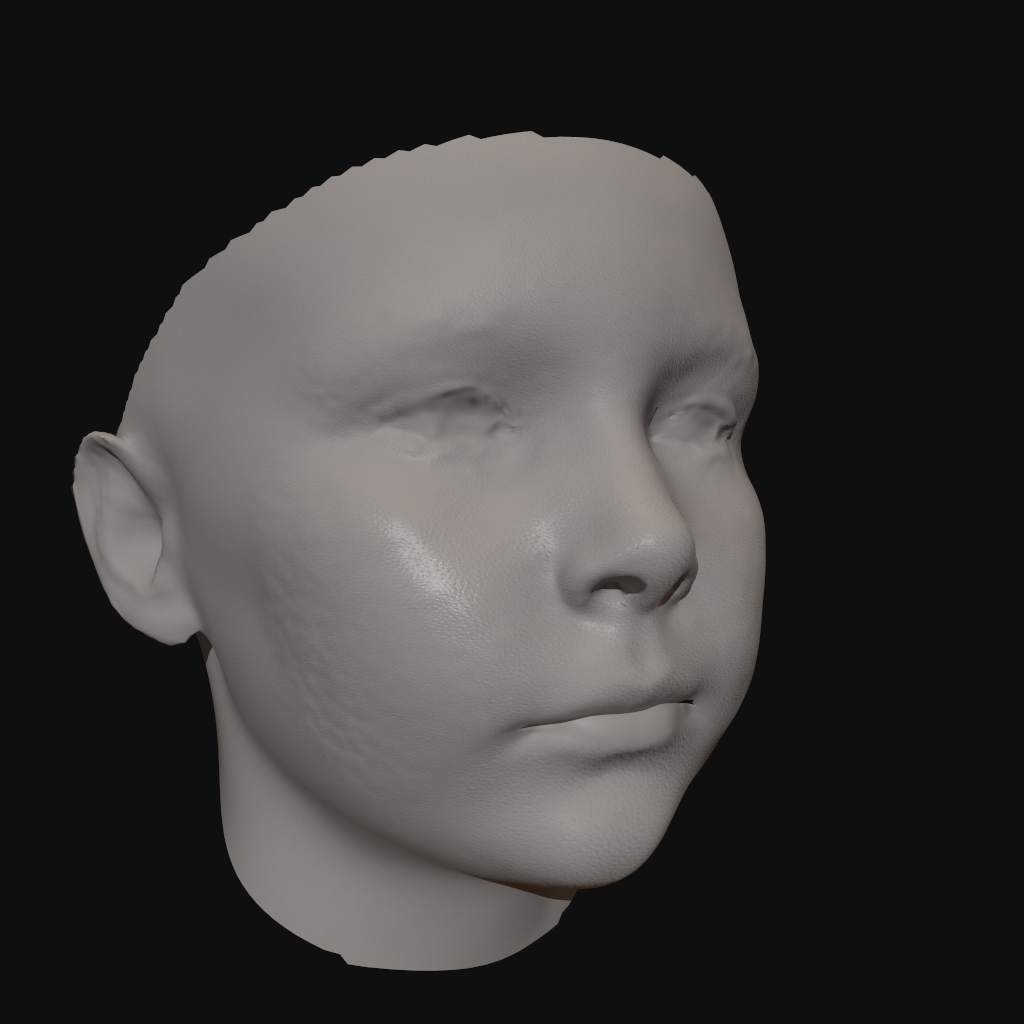}
\caption{Shp+Nor}\end{subfigure}
\begin{subfigure}{\var\textwidth}
	\includegraphics[width=1.0\textwidth]{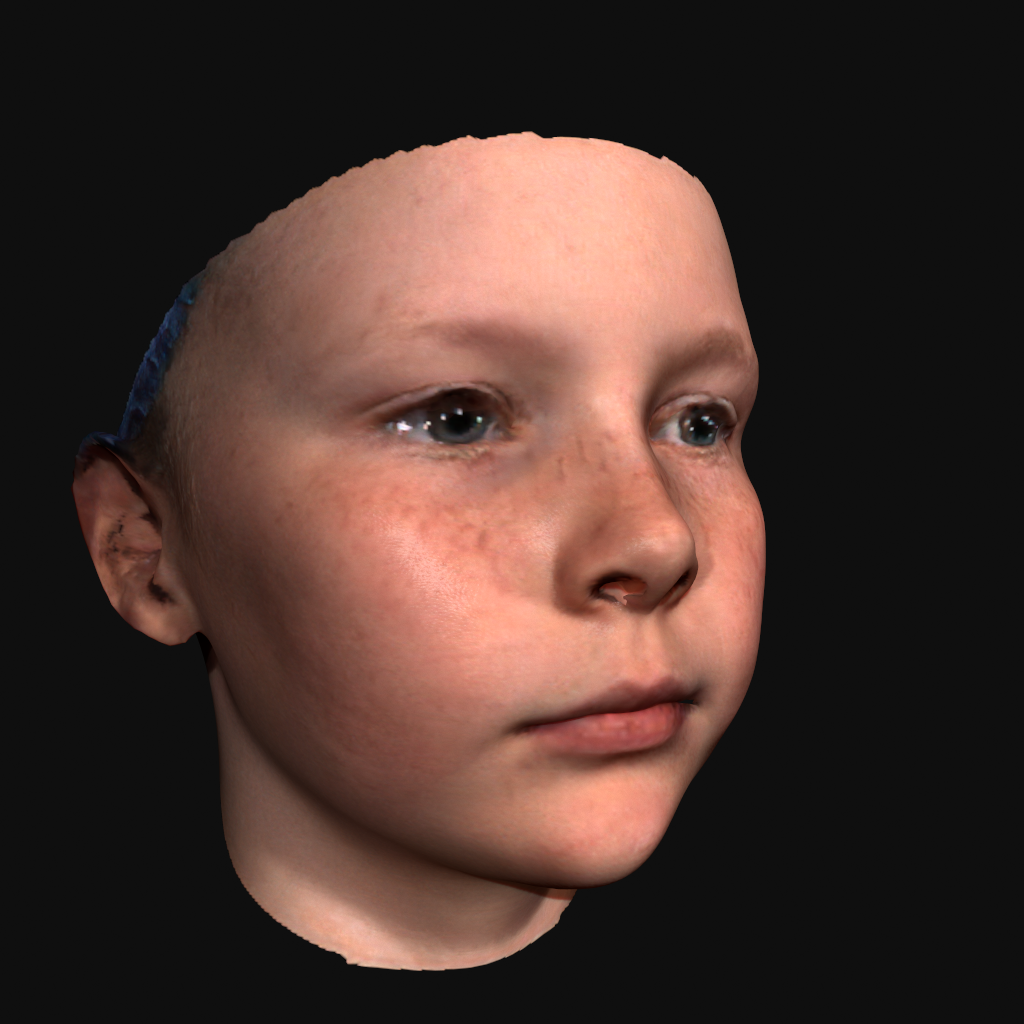}
\caption{All}\end{subfigure}
\caption{Generated UV representations and their corresponding additive renderings. Please note the strong correlation between UV maps, high fidelity and photorealistic renderings. The figure is best viewed in zoom.}
\label{fig:uvmaps}
\end{figure*}

\subsubsection{Diversity:} Our model is well-generalized with different age, gender, ethnicity groups and many facial attributes. Although Fig.~\ref{fig:diversity} shows diversity in some of those categories, the reader is encouraged to see identity variation throughout the paper and the supplementary video. 

\def \var {0.196}
\begin{figure*}[t!]
\centering
\begin{subfigure}{\var\textwidth}
\includegraphics[width=0.5\textwidth]{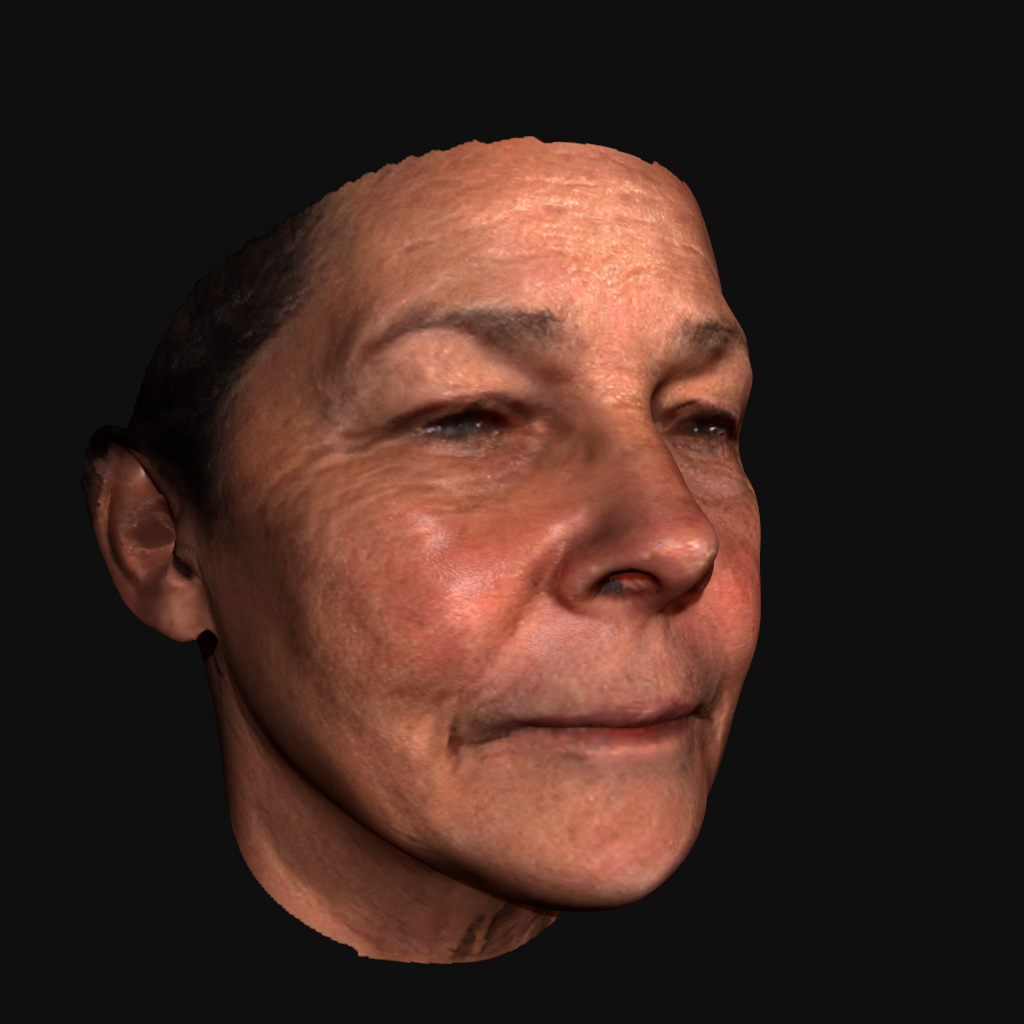}\includegraphics[width=0.5\textwidth]{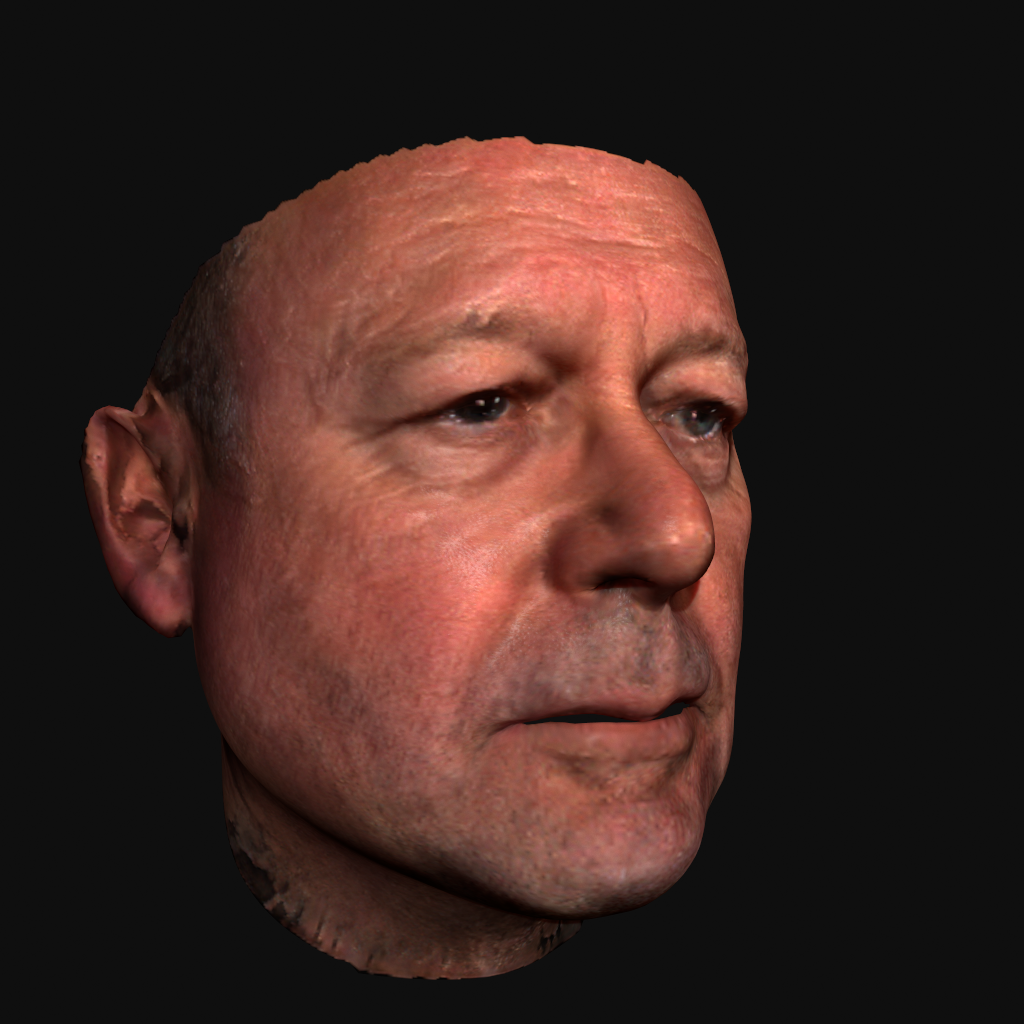}
\includegraphics[width=0.5\textwidth]{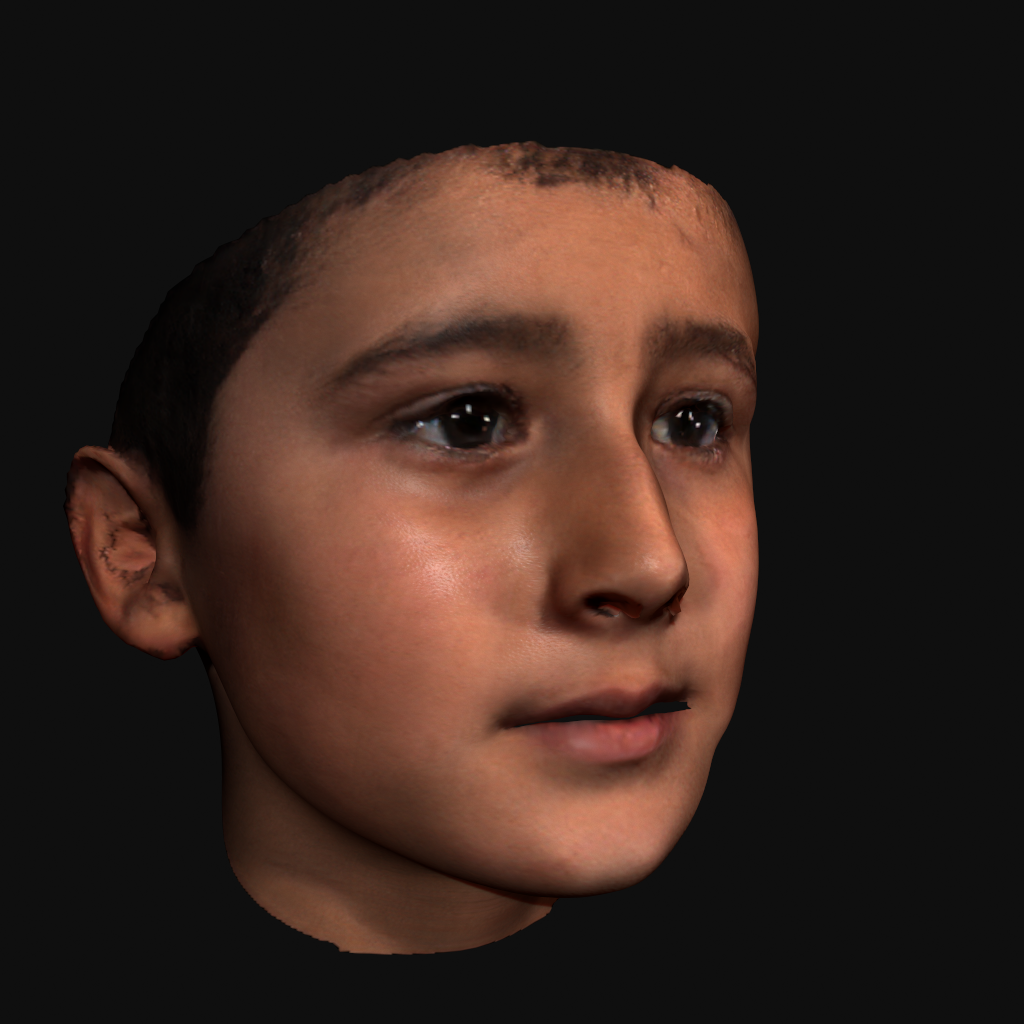}\includegraphics[width=0.5\textwidth]{qualitative/327_shp_tex_nor_0_30_60/000041_30}
\tiny\caption{Age}\end{subfigure}\hfill\begin{subfigure}{\var\textwidth}
{\includegraphics[width=0.5\textwidth]{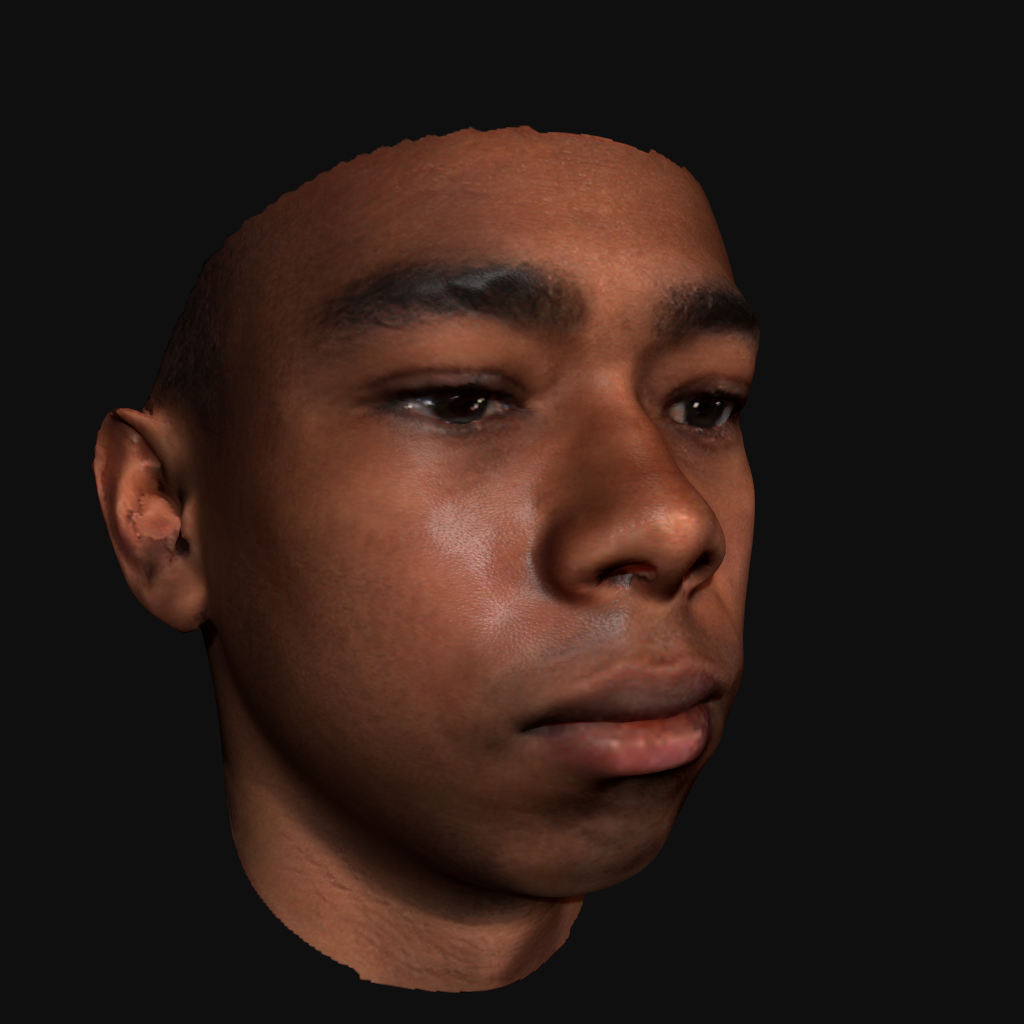}\includegraphics[width=0.5\textwidth]{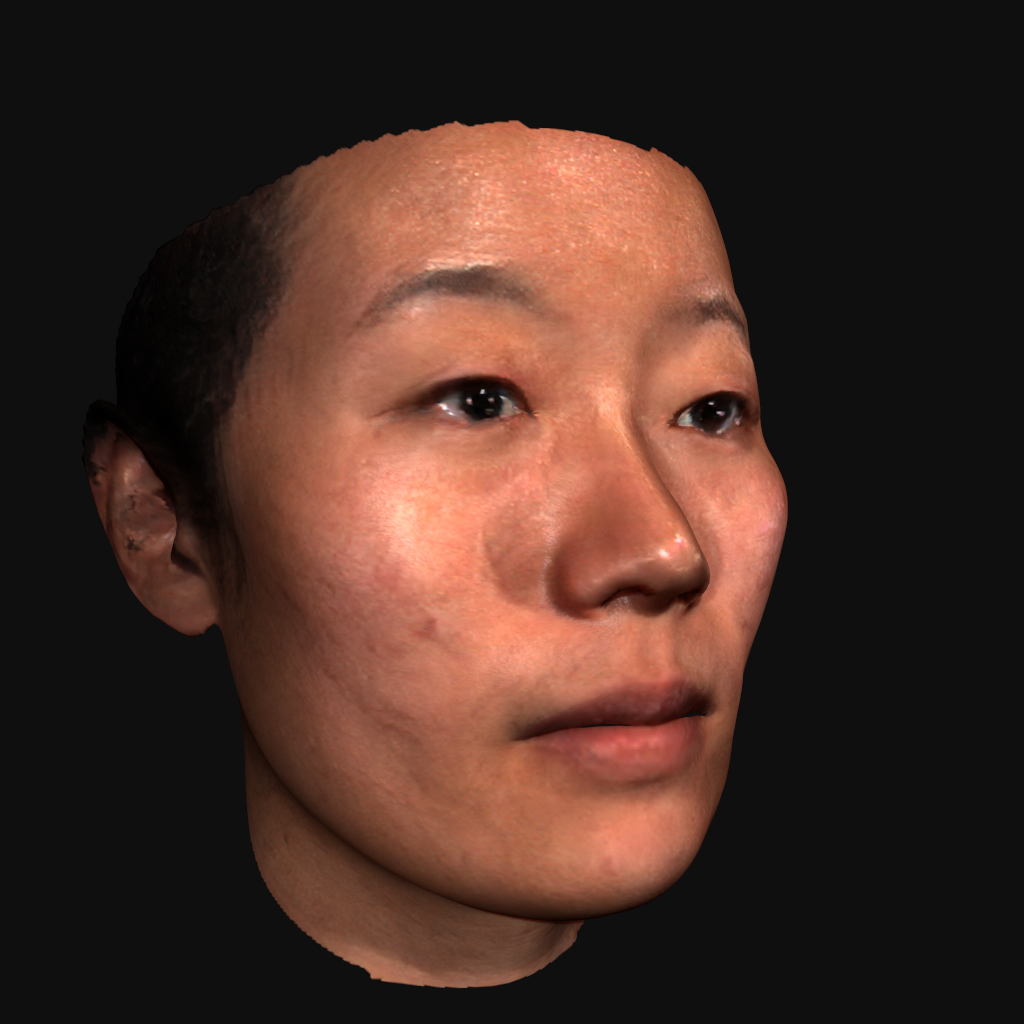}
\includegraphics[width=0.5\textwidth]{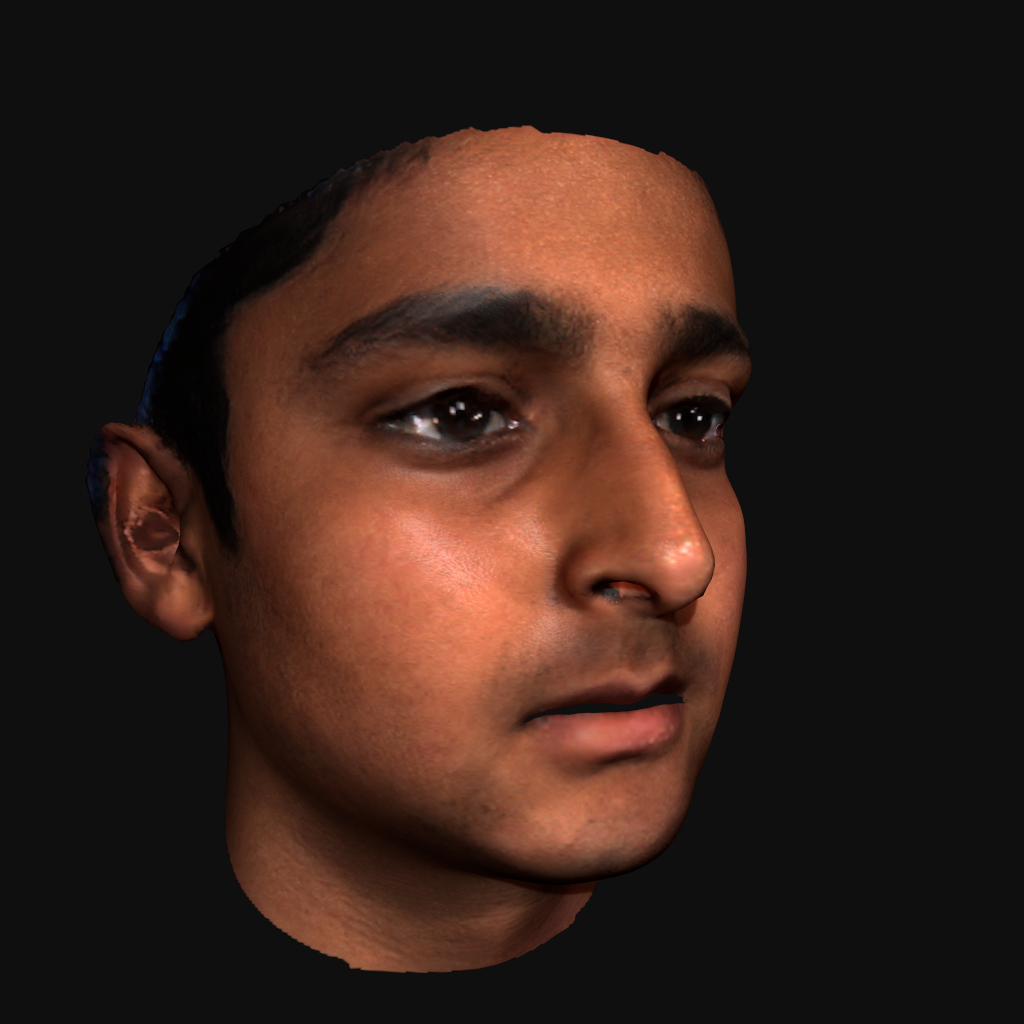}\includegraphics[width=0.5\textwidth]{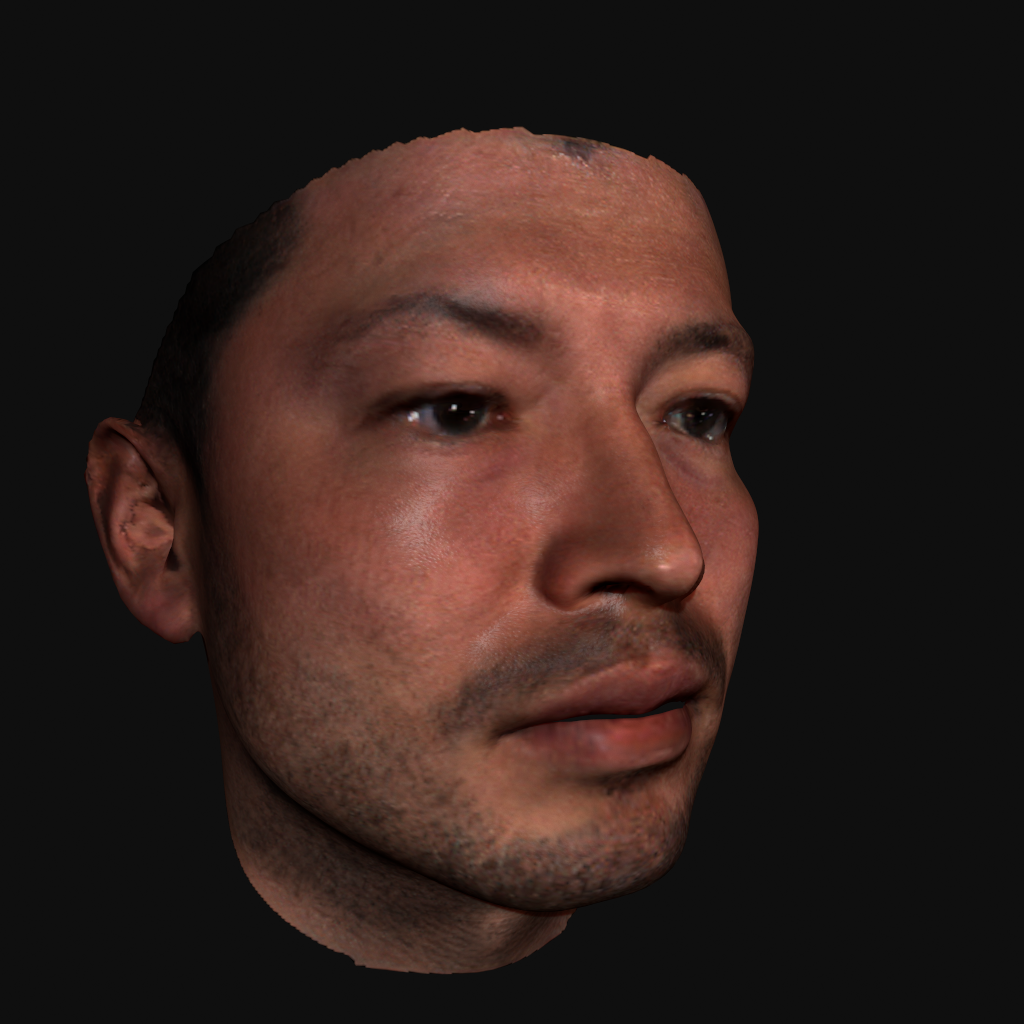}}
\tiny\caption{Ethnicity}\end{subfigure}\hfill\begin{subfigure}{\var\textwidth}
\includegraphics[width=0.5\textwidth]{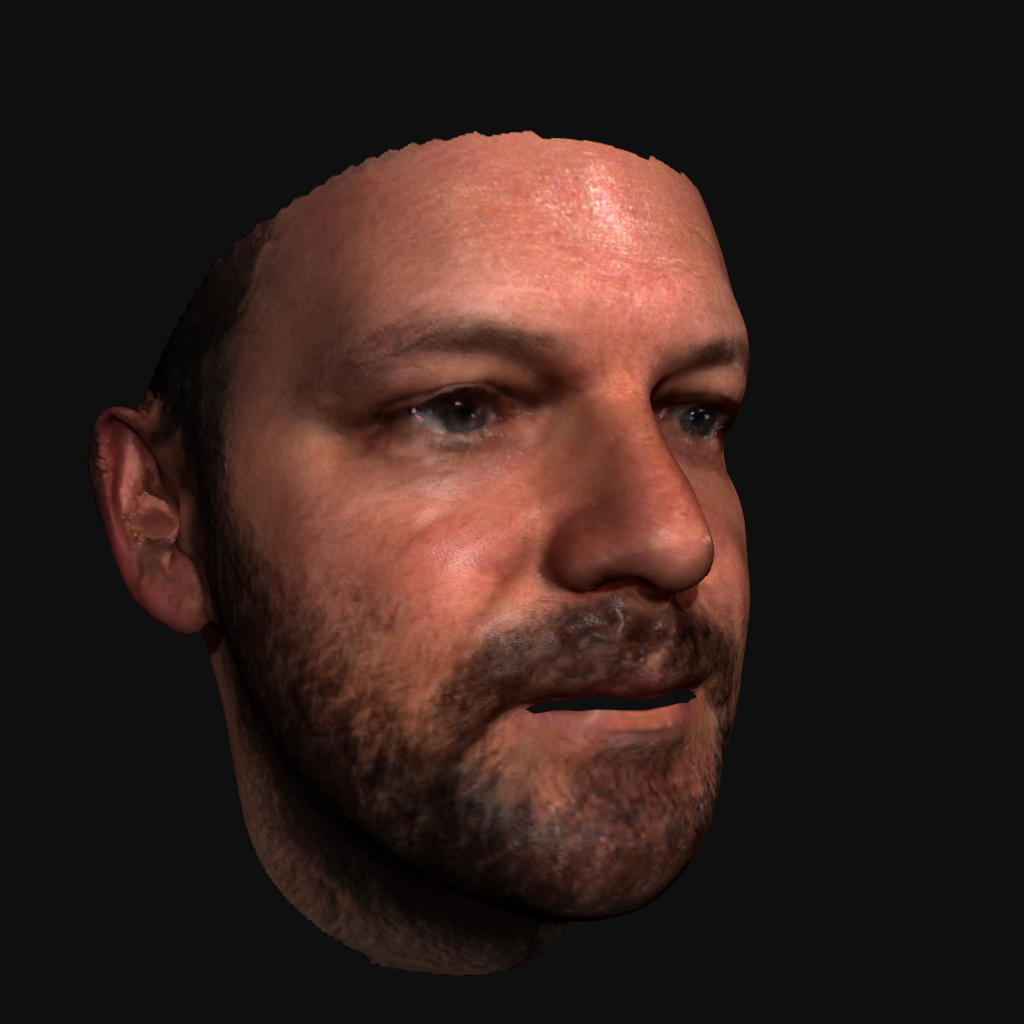}\includegraphics[width=0.5\textwidth]{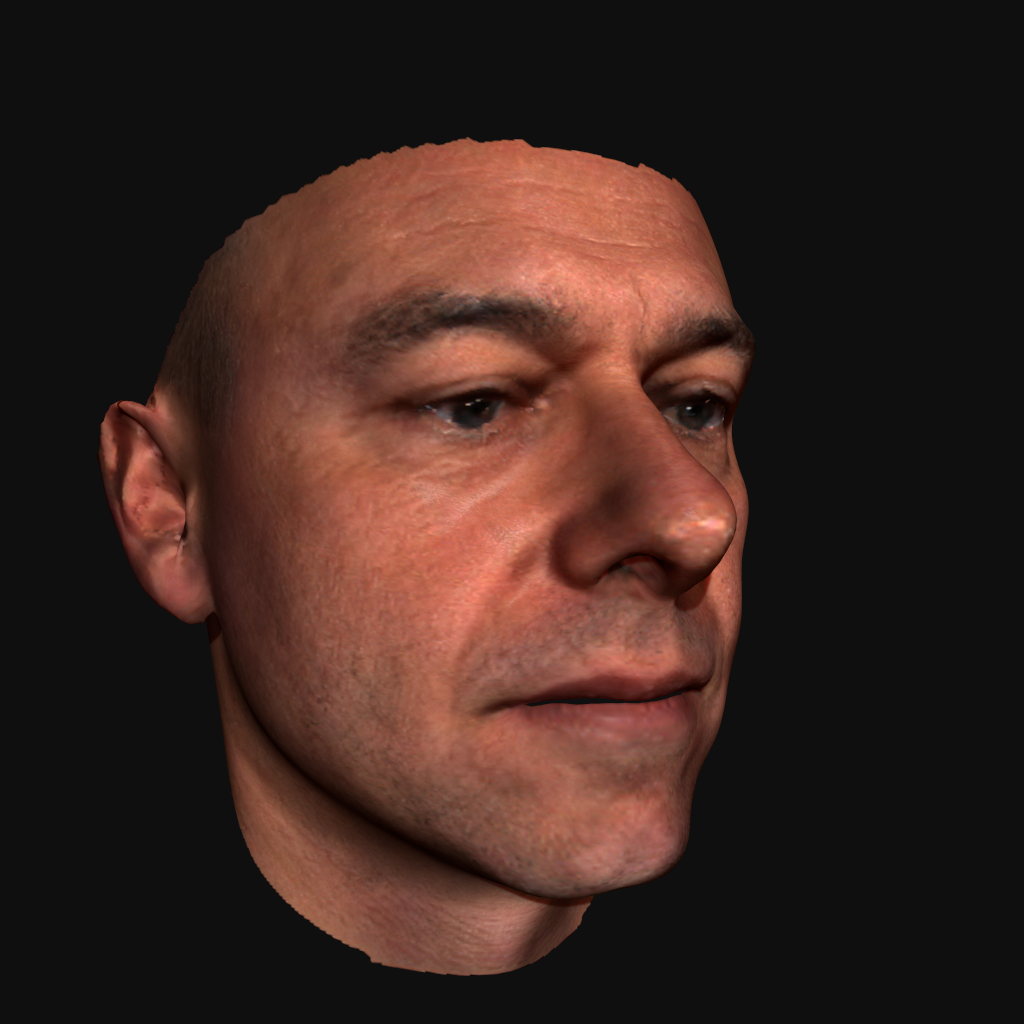}
\includegraphics[width=0.5\textwidth]{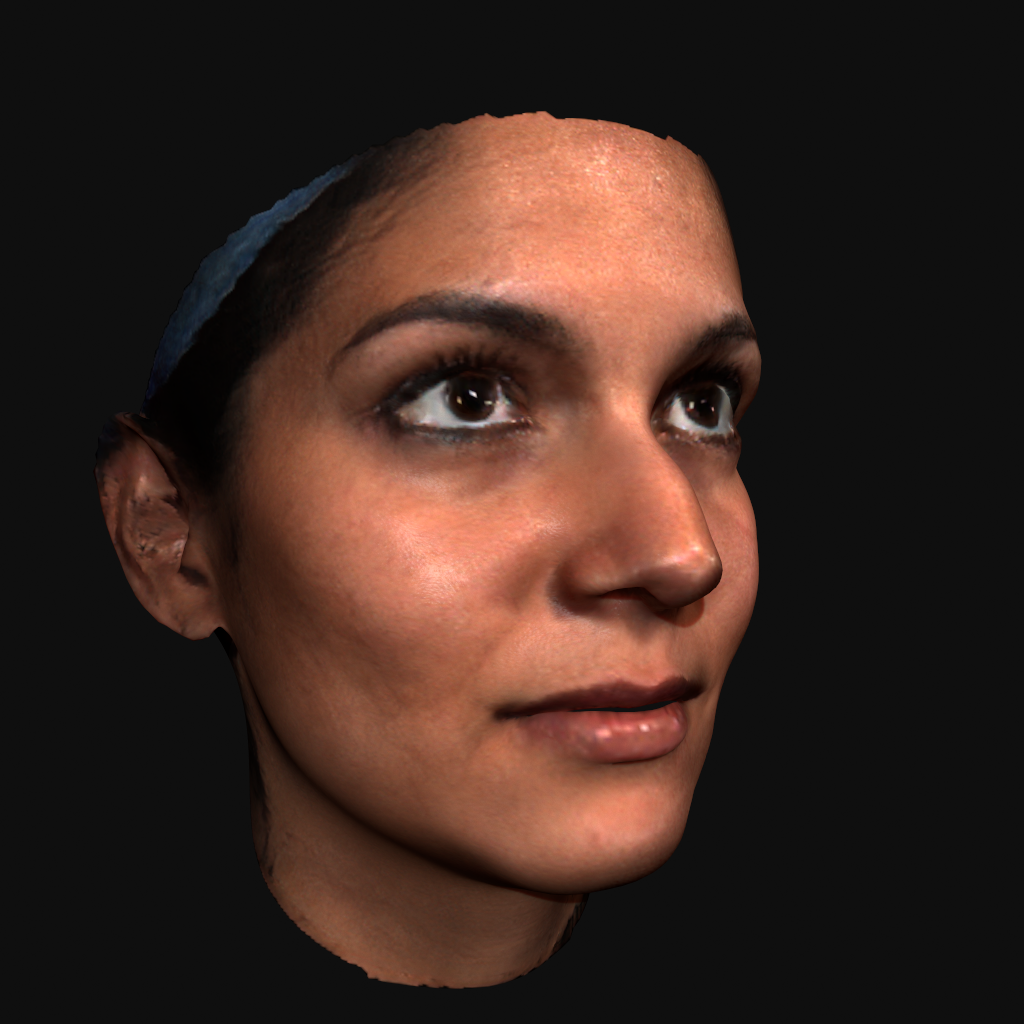}\includegraphics[width=0.5\textwidth]{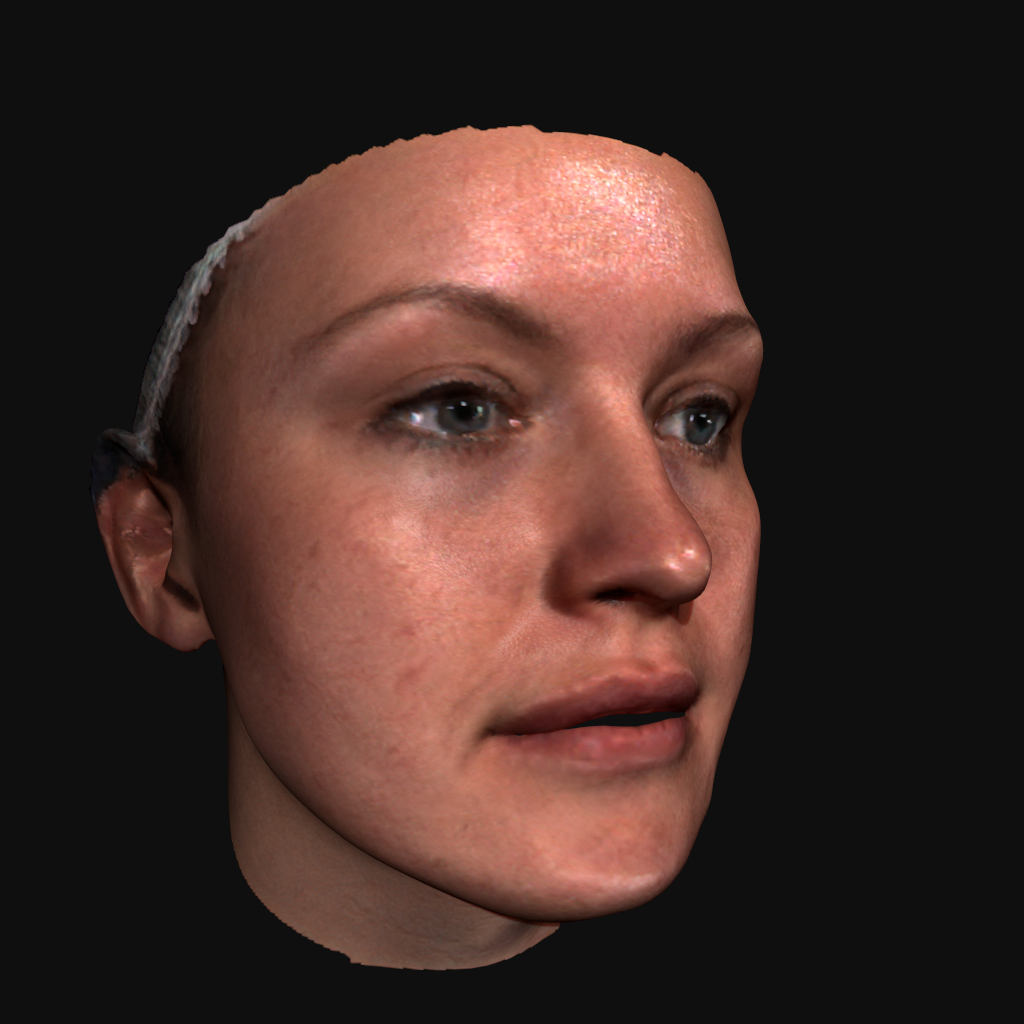}
\tiny\caption{Gender}\end{subfigure}\hfill
\begin{subfigure}{\var\textwidth}
\includegraphics[width=0.5\textwidth]{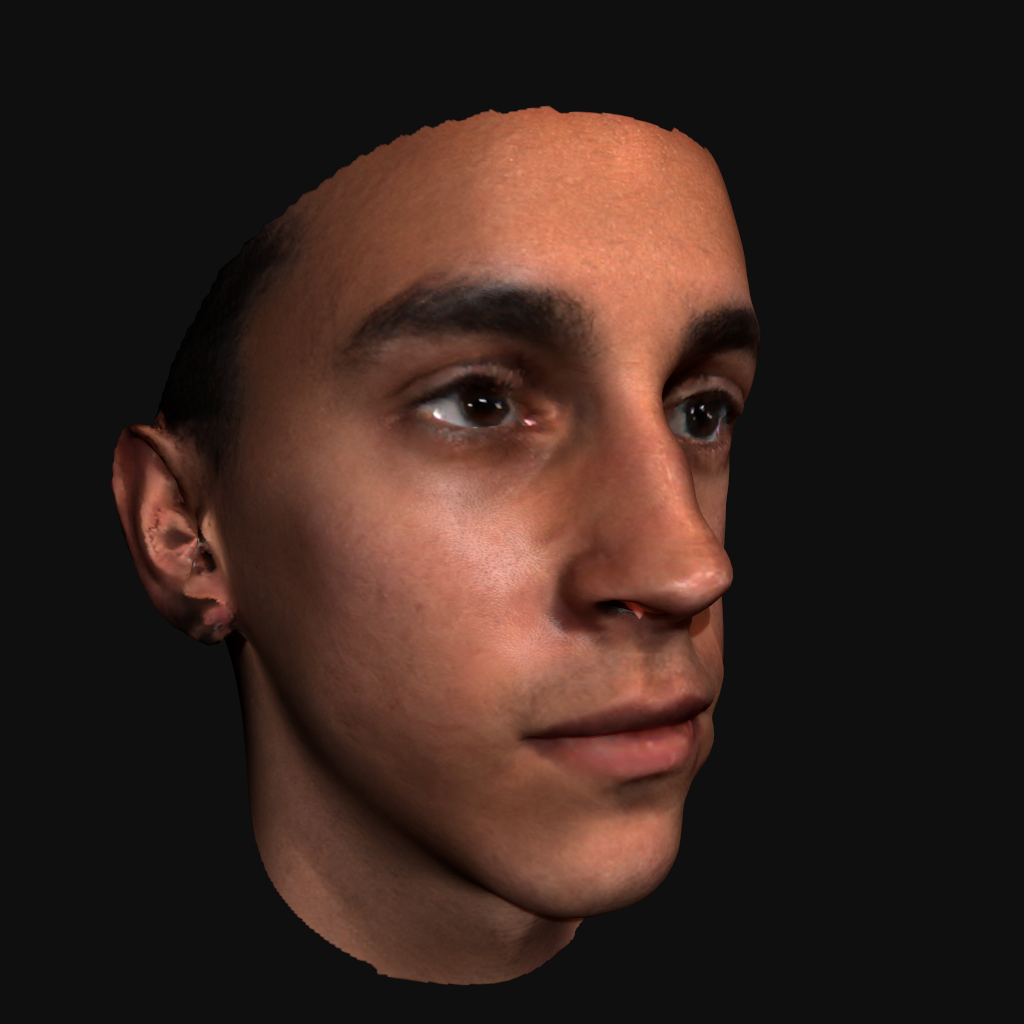}\includegraphics[width=0.5\textwidth]{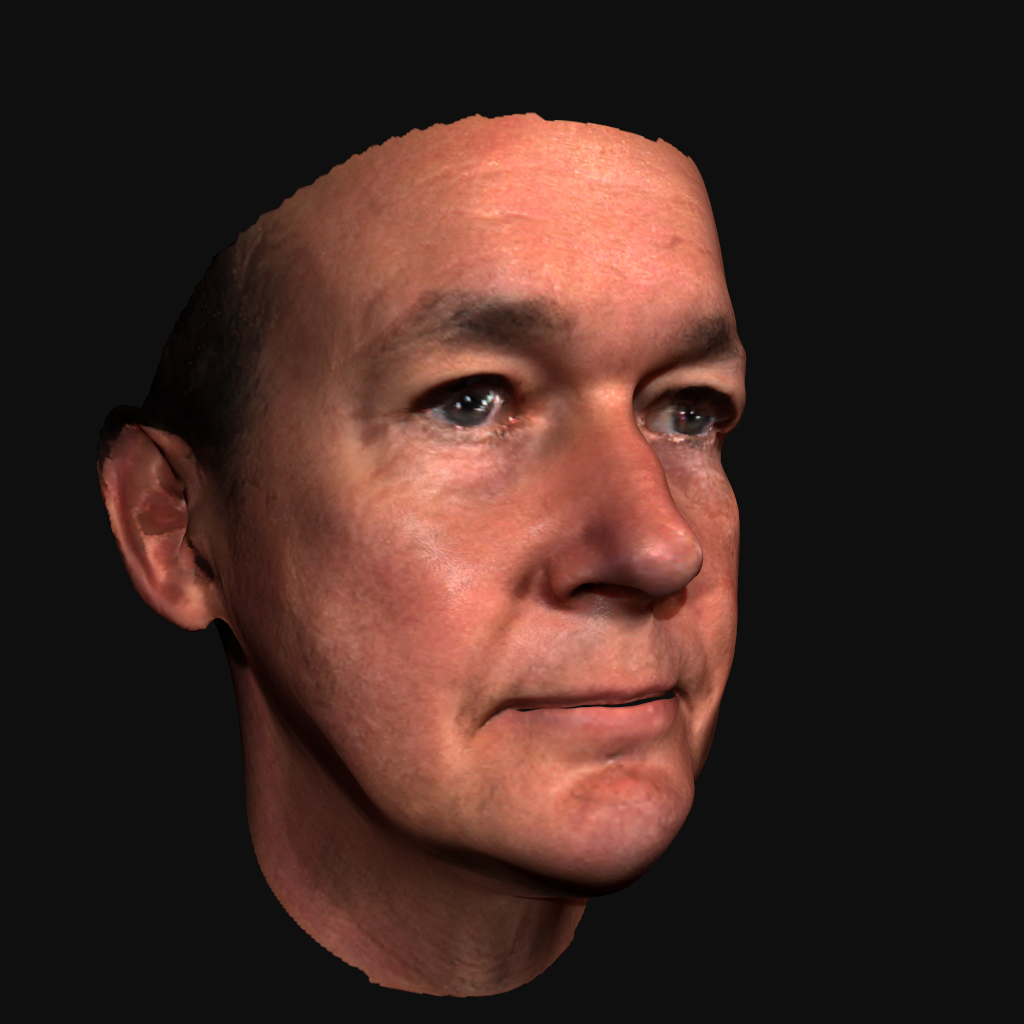}
\includegraphics[width=0.5\textwidth]{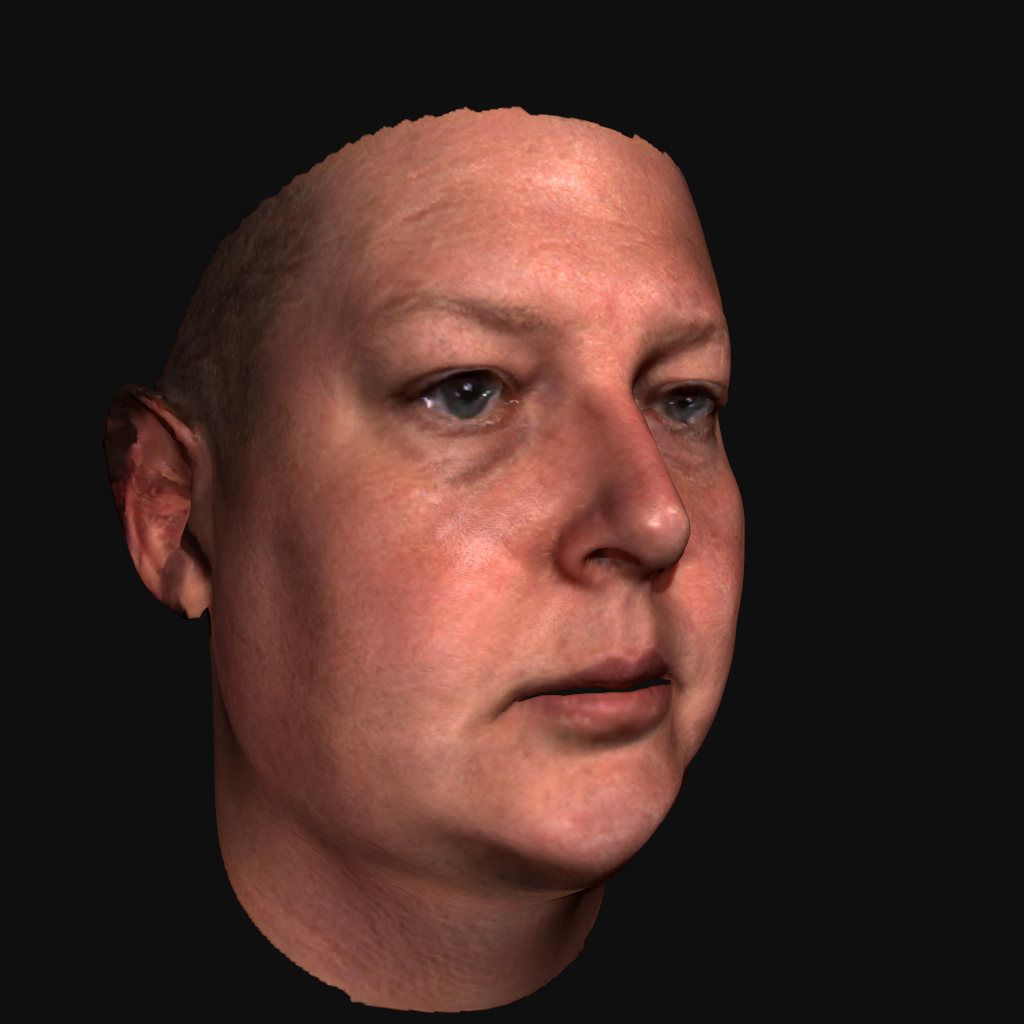}\includegraphics[width=0.5\textwidth]{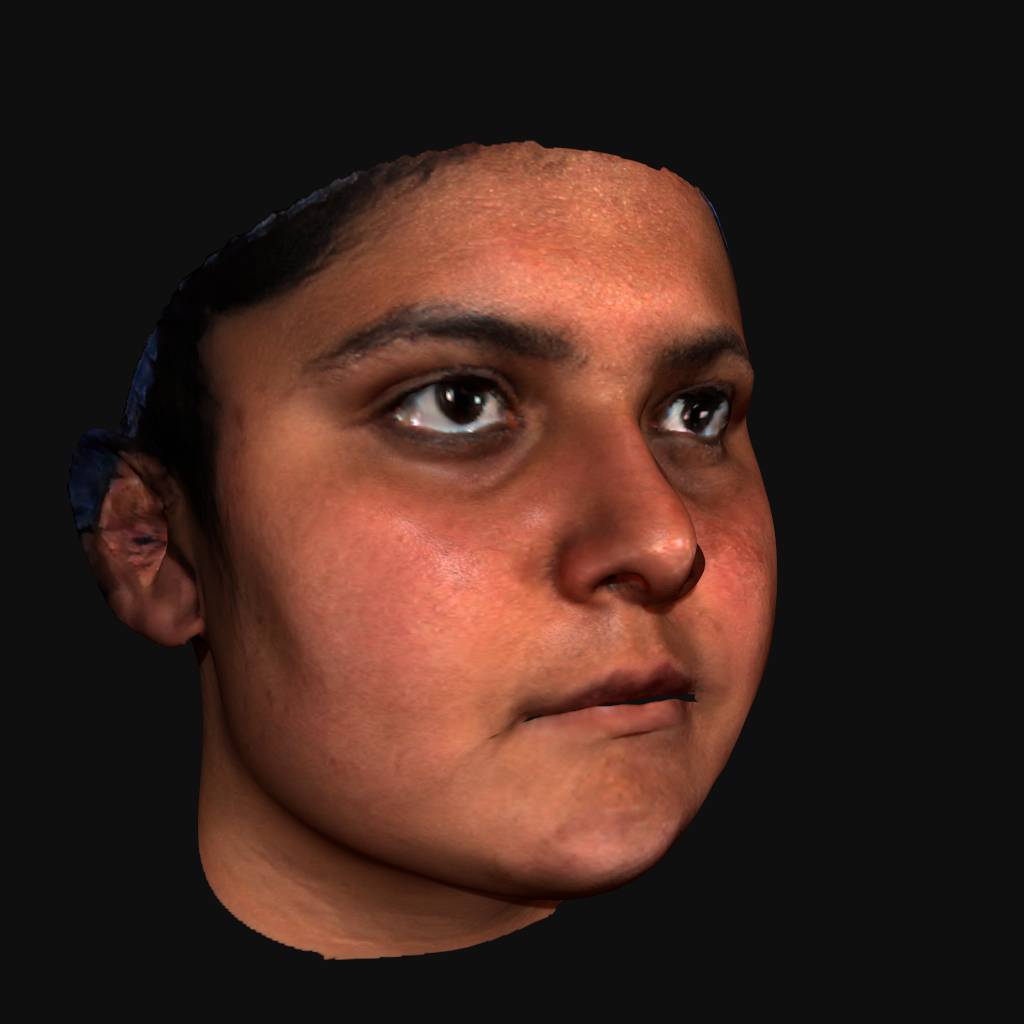}
\tiny\caption{Weight}\end{subfigure}\hfill
\begin{subfigure}{\var\textwidth}
\includegraphics[width=0.5\textwidth]{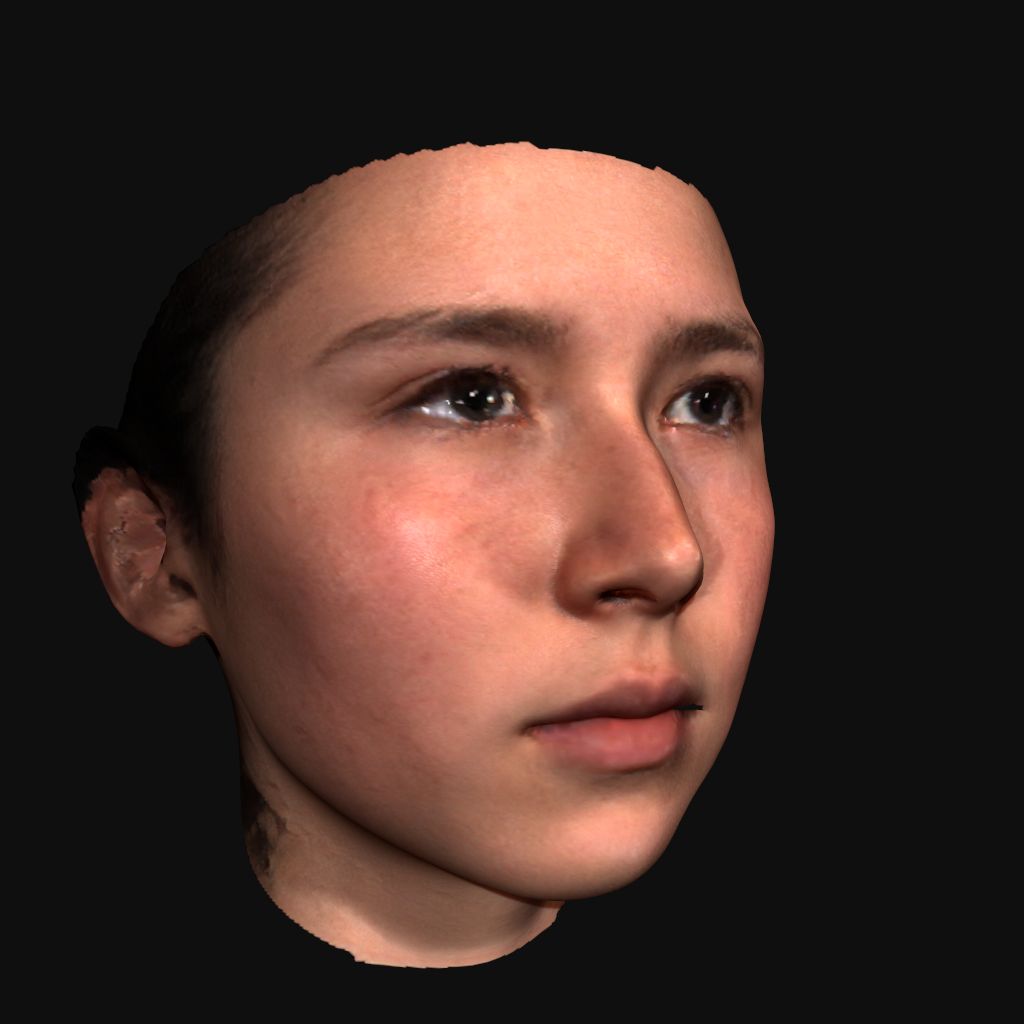}\includegraphics[width=0.5\textwidth]{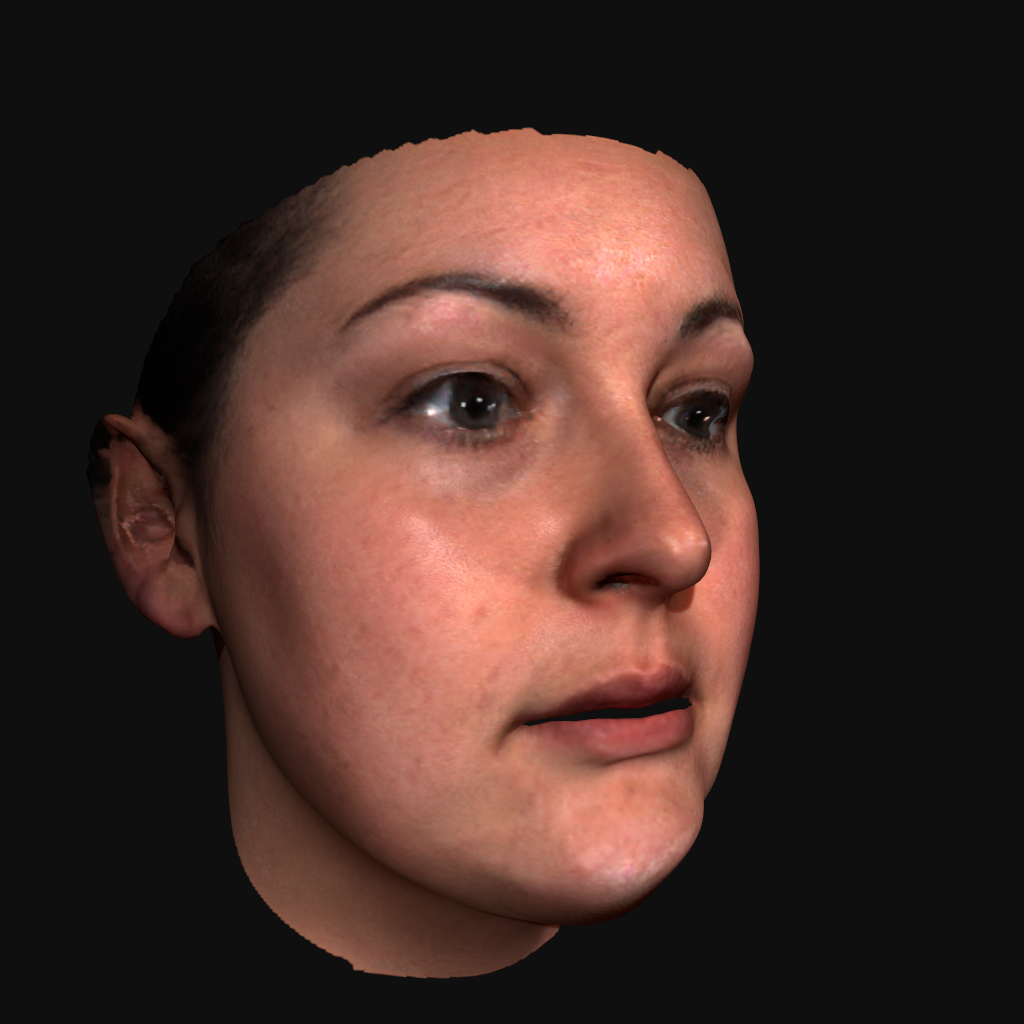}
\includegraphics[width=0.5\textwidth]{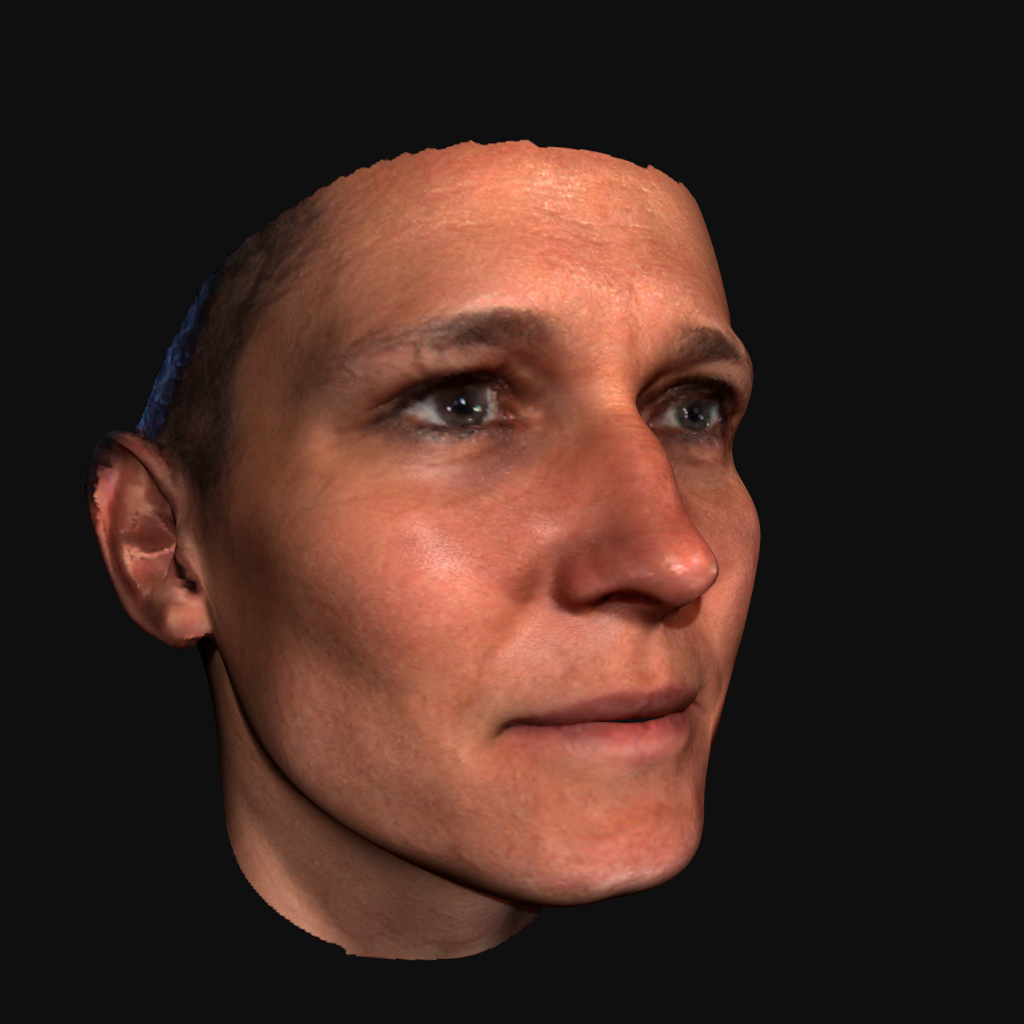}\includegraphics[width=0.5\textwidth]{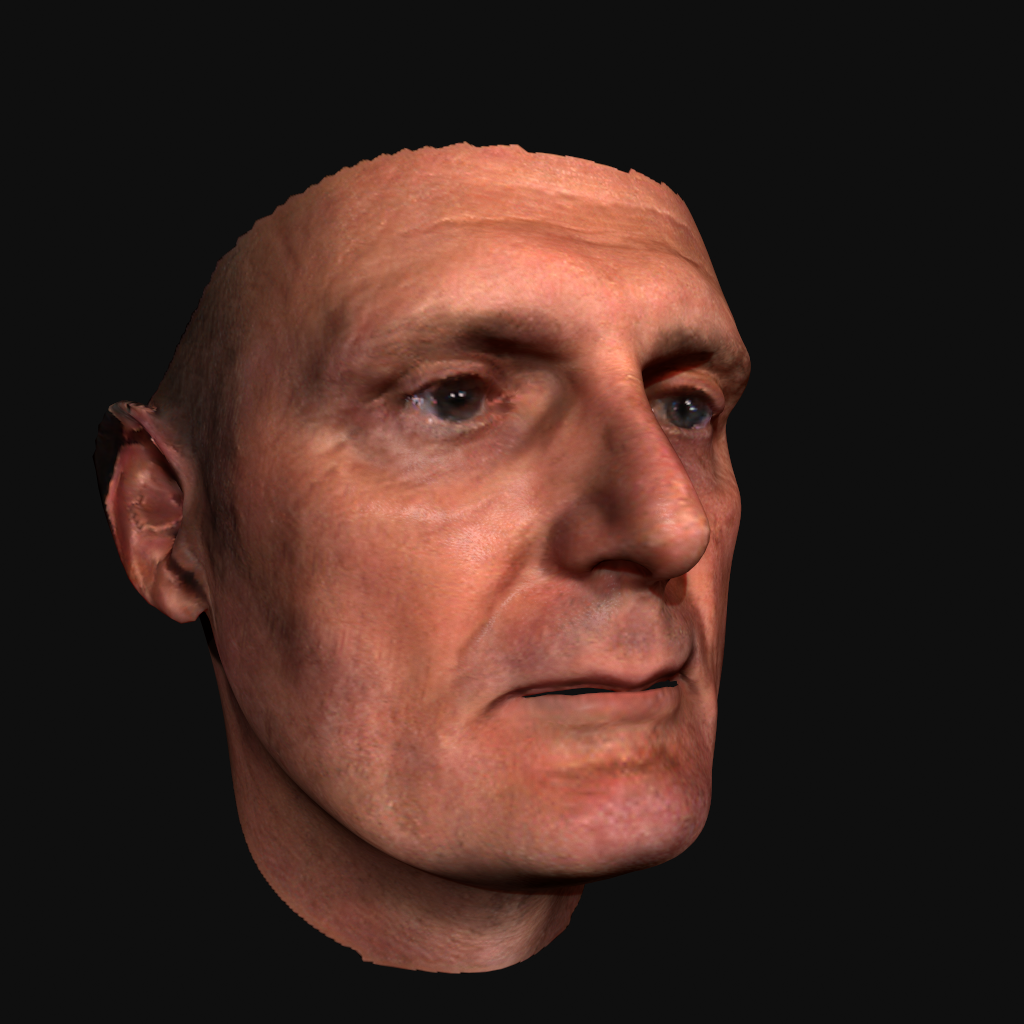}
\tiny\caption{Roundness}\end{subfigure}\hfill

%
\caption{Variation of generated 3D faces by our model. Each block shows diversity in a different aspect. Readers are encouraged to zoom in on a digital version.}
\label{fig:diversity}
\end{figure*}

\subsubsection{Expression:} We also show that our expression generator is capable of synthesizing quite a diverse set of expressions. Moreover, the expressions can be controlled by the input label as can be seen in Fig.~\ref{fig:exp}. The reader is encouraged to see more expression generations in the supplementary video.

\def \var {0.083}
\begin{figure}[t]
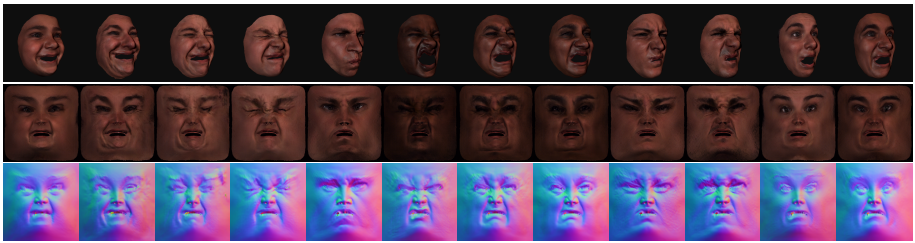

\centering
\ForEach
{,}
{\includegraphics[width=\var\textwidth]{exp/00\thislevelitem}}
{1364,0144,0324,1094,1524,1618,1004,0982,1153,1534,0043,0927}\\
\ForEach
{,}
{\includegraphics[width=\var\textwidth]{exp/00\thislevelitem_tex}}
{1364,0144,0324,1094,1524,1618,1004,0982,1153,1534,0043,0927}\\
\ForEach
{,}
{\includegraphics[width=\var\textwidth]{exp/00\thislevelitem_nor}}
{1364,0144,0324,1094,1524,1618,1004,0982,1153,1534,0043,0927}

\caption{(Top) generations of six universal expressions (i.e. each two columns respective the following expressions: Happiness, Sadness, Anger, Fear, Disgust, Surprise). (Middle) texture and (Bottom) normals maps are used to generate the corresponding 3D faces. Please note how expressions are represented and correlated in the texture and normals space.}
\label{fig:exp}
\end{figure}

\subsubsection{Interpolation between identities:} As shown in the supplementary video and in Fig.~\ref{fig:interpolate}, our model can easily interpolate between any generation in a visually continuous set of identities which is another indication that the model is free from mode collapse. Interpolation is done by randomly generating two identities and generates faces by evenly spaced samples in latent space between the two.

\subsubsection{Full head completion:}

\def \var {0.1}
\begin{figure*}[t]
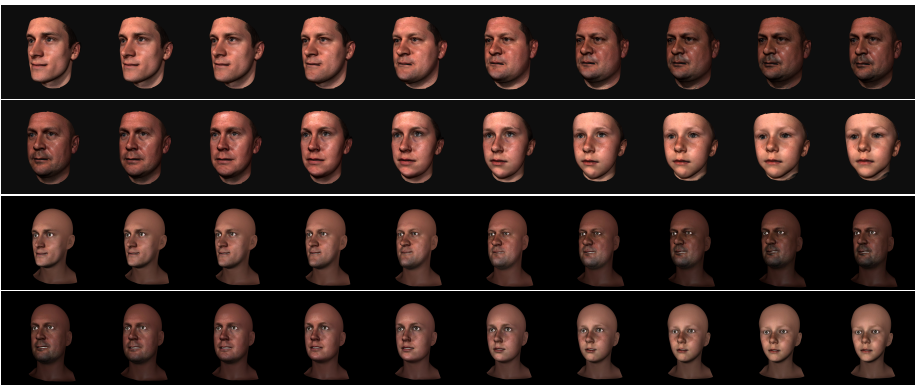

\centering
\ForEach
{,}
{\includegraphics[width=\var\textwidth]{interpolation/face/00\thislevelitem0}}{118,119,120,121,122,123,124,125,126,127}
\ForEach
{,}
{\includegraphics[width=\var\textwidth]{interpolation/face/00\thislevelitem0}}{128,129,130,131,132,133,134,135,136,137}
\ForEach
{,}
{\includegraphics[width=\var\textwidth]{interpolation/head/00\thislevelitem0}}{118,119,120,121,122,123,124,125,126,127}
\ForEach
{,}
{\includegraphics[width=\var\textwidth]{interpolation/head/00\thislevelitem0}}{128,129,130,131,132,133,134,135,136,137}
\caption{Interpolation between pair of identities in the latent space. Smooth transition indicates generalization of our GAN model. The last two rows show complete full head representations respective to the first two rows.}
\label{fig:interpolate}
\end{figure*}

%
%
%
%

We also extend our facial 3D meshes to full head representations by employing the framework proposed in \cite{ploumpis2019combining}. We achieve this by regressing from a latent space that represents only the 3D face to the PCA latent space of the Universal Head Model (UHM)~\cite{ploumpis2019combining,ploumpis2020complete}. We begin by building a PCA model of the inner face based on the $10,000$ neutral scans of the MeIn3D dataset. Similarly, we exploit the extended full head meshes of the same identities utilized by UHM model and project them to the UHM subspace to acquire the latent shape parameters of the entire head topology. Finally, we learn a regression matrix by solving a linear least-square optimization problem as proposed in \cite{ploumpis2019combining}, which maps the latent space of the face shape to the full head representation. Fig.~\ref{fig:interpolate} demonstrates the extended head representations of our approach in conjunction with the synthesized crop faces.

\subsubsection{Comparison to decoupled modalities and PCA:}
\begin{figure}[t]
\centering
\includegraphics[width=0.8\linewidth]{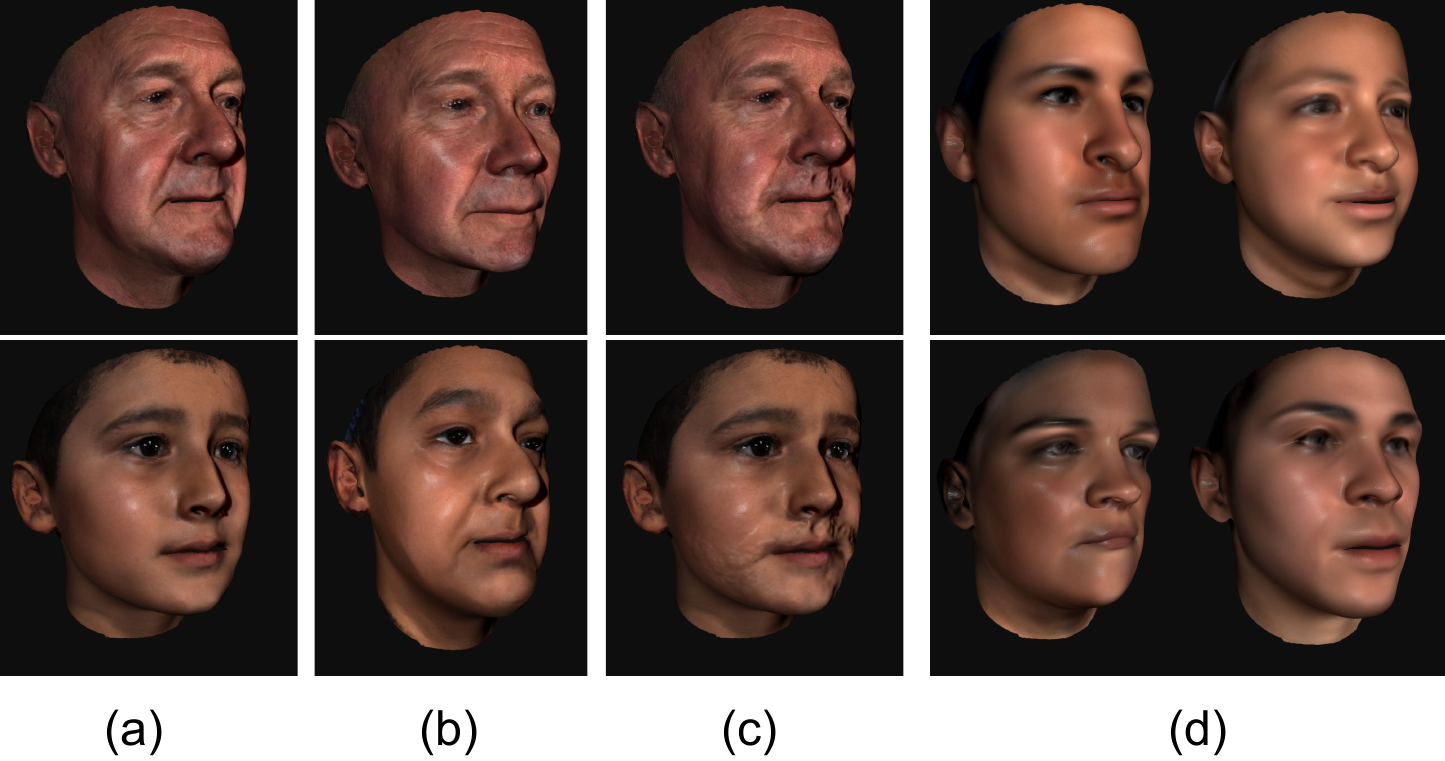}
\caption{Comparison with seperate GAN models and PCA model. (a) Generation by our model. (b) Same texture with random shape and normals. (c) Same texture and shape with random normals (i.e. beard). (d) Generation by a PCA model constructed by the same training data and the same identity-generic rendering tools as explained in Sec.~3.4.}
\label{fig:mixed}
\end{figure}

Results in Fig.~\ref{fig:mixed} reveal a set of advantages of such unified 3D face modeling over separate GAN and statistical models. Clearly, the figure shows that the correlation among texture, shape, and normals is an important component for realistic face synthesis. Also, generations by PCA models are missing photorealism and details significantly.

\subsection{Pose-invariant Face Recognition}

In this section, we present an experiment that demonstrates that the proposed methodology can generate faces of different and diverse identities. That is, we use the generated faces to train one of the most recent state-of-the-art face recognition method, ArcFace~\cite{deng2018arcface}, and show that the proposed shape and texture generation model can boost the performance of pose-invariant face recognition.

\begin{figure}[t]
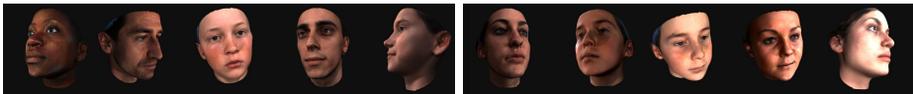

\centering
\ForEach
{,}
{\includegraphics[width=0.099\textwidth]{random/\thislevelitem}}{000000_6,000002_20,000005_14,000035_39,000099_42}
\ForEach
{,}
{\includegraphics[width=0.099\textwidth]{random/\thislevelitem}}{000009_10,000011_43,000011_44,000019_30,000030_5}
\caption{Examples of generated data (``Gen'') by the proposed method.}
\label{fig:face-recognition}
\end{figure}

\noindent{\bf Training Data:} We randomly synthesize 10K new identities from the proposed model and render 50 images per identity with a random camera and illumination parameters from the Gaussian distribution of the 300W-LP dataset~\cite{zhu2016face,gecer2018facegan}. For clarity, we call this dataset ``Gen'' in the rest of the text. Fig.~\ref{fig:face-recognition} illustrates some examples of ``Gen'' dataset which show larger pose variations than the real-world collected data. We augment ``Gen'' with an in-the-wild training data, CASIA dataset~\cite{yi2014learning}, which consists of 10,575 identities with 494,414 images.

\noindent{\bf Test Data:} For evaluation, we employ Celebrities in Frontal Profile (CFP)~\cite{sengupta2016frontal} and Age Database (AgeDB)~\cite{Moschoglou2017AgeDB}. {\bf CFP}~\cite{sengupta2016frontal} consists of 500 subjects, each with 10 frontal and 4 profile images. The evaluation protocol includes frontal-frontal (FF) and frontal-profile (FP) face verification. In this paper, we focus on the most challenging subset, CFP-FP, to investigate the performance of pose-invariant face recognition. There are 3,500 same-person pairs and 3,500 different-person pairs in CFP-FP for the verification test. {\bf AgeDB}~\cite{Moschoglou2017AgeDB,deng2017marginal} contains $12,240$ images of $440$ distinct subjects. The minimum and maximum ages are $3$ and $101$, respectively. The average age range for each subject is $49$ years. There are four groups of test data with different year gaps ($5$ years, $10$ years, $20$ years and $30$ years, respectively)~\cite{deng2017marginal}. In this paper, we only use the most challenging subset, AgeDB-30, to report the performance. There are 3,000 positive pairs and 3,000 negative pairs in AgeDB-30 for the verification test. 

\noindent{\bf Data Preprocessing:} We follow the baseline~\cite{deng2018arcface} to generate the normalized face crops ($112\times112$) by utilizing five facial points. 

\noindent{\bf Training and testing Details:} For the embedding networks, we employ the widely used ResNet50 architecture~\cite{he2016deep}. After the last convolutional layer, we also use the BN-Dropout-FC-BN~\cite{deng2018arcface} structure to get the final $512$-$D$ embedding feature. For the hyper-parameter setting and loss functions, we follow~\cite{deng2018arcface,gecer2017learning,deng2017marginal}. The overlapping identities between the CASIA data set and the test set are removed for strict evaluations, and we only use a single crop for all testing.

\begin{figure}[t]
\begin{floatrow}
\capbtabbox{\begin{tabular}{l|ccc}
\hline
Methods                   & CFP-FP      & AgeDB-30\\
\hline
UVGAN~\cite{deng2018uv}   & 94.05      &   94.18\\
\hline
Ours (CASIA)       & 95.56       & 95.15\\
Ours (CASIA+Gen)  & {\bf 97.12}       & {\bf 95.18}\\
\hline
\end{tabular}
}{%
\caption{Verification performance ($\%$) of different models on CFP-FP and AgeDB-30.}
\label{tab:arcfacevalidation}
}
\ffigbox{
\begin{subfigure}{0.23\textwidth}
	\includegraphics[width=1.0\textwidth]{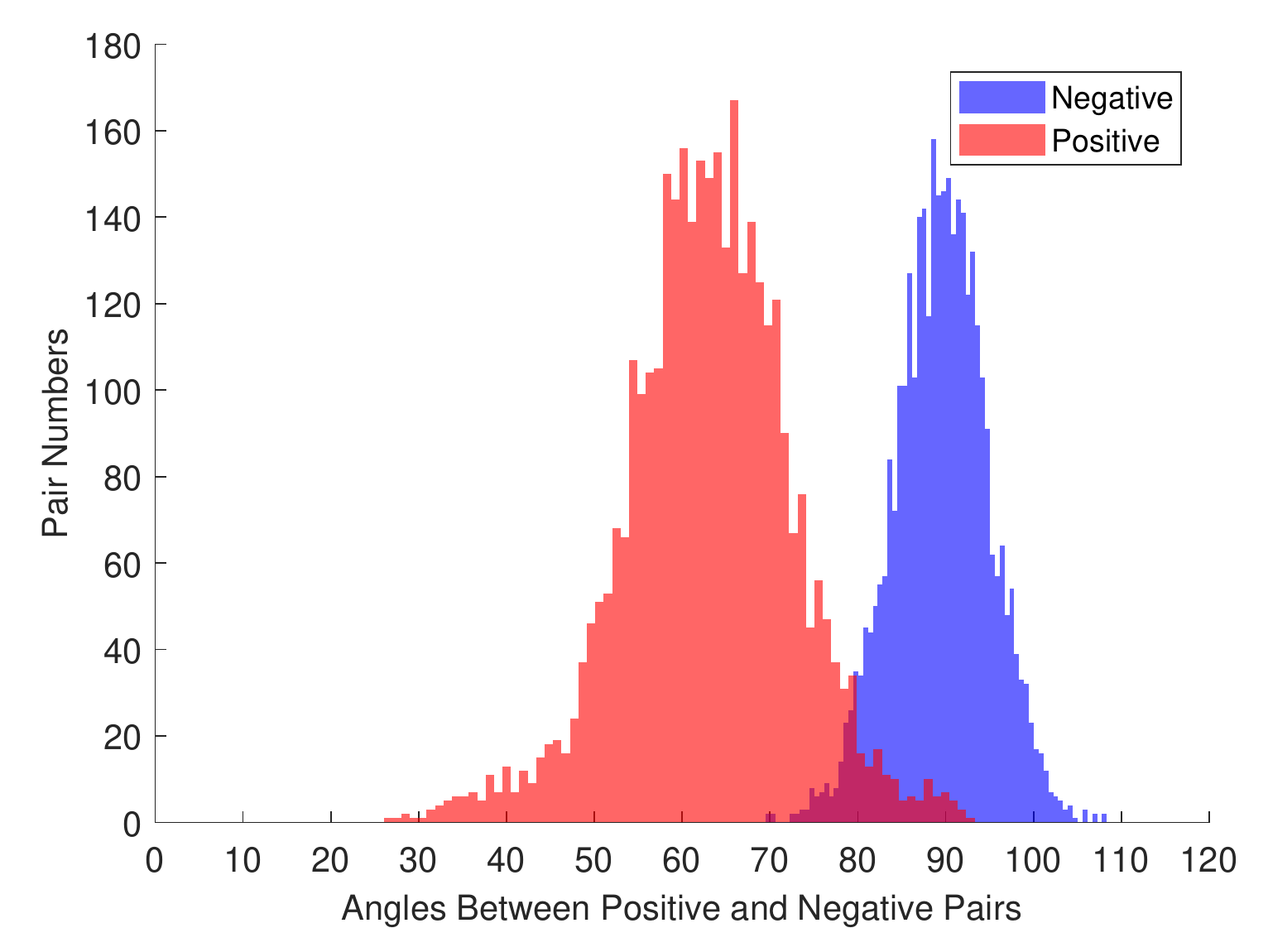}
\caption{\tiny{CASIA}}\label{fig:arc}\end{subfigure}
\begin{subfigure}{0.23\textwidth}
	\includegraphics[width=1.0\textwidth]{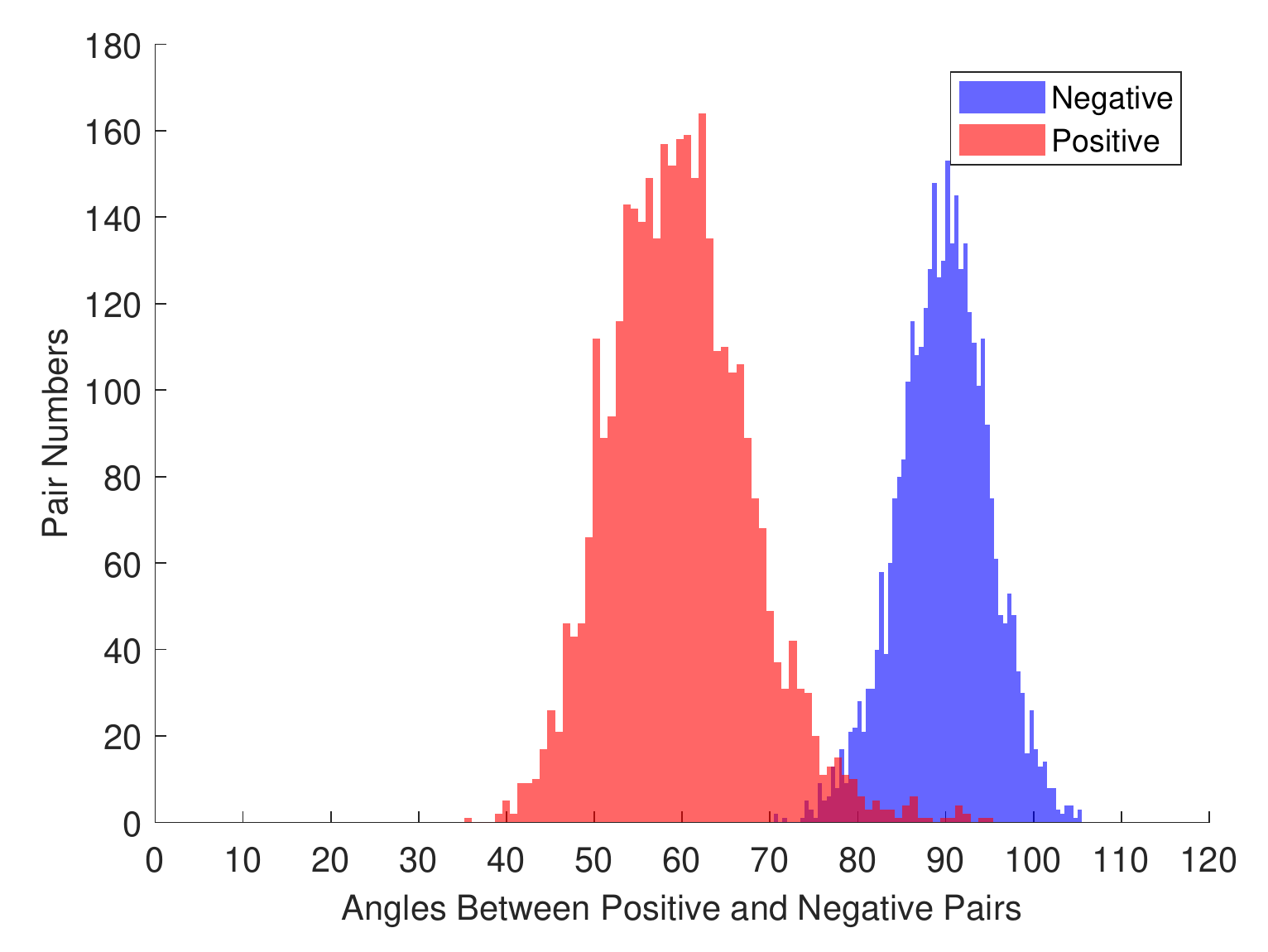}
\caption{\tiny{CASIA+Gen}}\label{fig:arcgen2}\end{subfigure}
}{%
\caption{Angle distributions of CFP-FP  positive (red) and negative (blue) pairs in the $512$-$D$ feature space. 
}
\label{fig:histpairs}
}
\end{floatrow}
\end{figure}

\noindent{\bf Result Analysis:} In Table~\ref{tab:arcfacevalidation}, we show the contribution of the generated data on pose-invariant face recognition. We take UV-GAN~\cite{deng2018uv} as the baseline method, which attaches the completed UV texture map onto the fitted mesh and generates instances of arbitrary poses to increase pose variation during training and minimize pose discrepancy during testing. As we can see from Table~\ref{tab:arcfacevalidation}, generated data significantly boost the verification performance on CFP-FP from $95.56\%$ to $97.12\%$, decreasing the verification error by $51.2\%$ compared to the result of UV-GAN~\cite{deng2018uv}. On AgeDB-30, combining CASIA and generated data achieves similar performance compared to using single CASIA because we only include intra-variance from pose instead of age. 


In Figure~\ref{fig:histpairs}, we show the angle distributions of all positive pairs and negative pairs from CFP-FP. By incorporating generation data, the overlap indistinguishable area between the positive histogram and the negative histogram is obviously decreased, which confirms that ArcFace can learn pose-invariant feature embedding from the generated data. In Table~\ref{tab:compactness}, we select some verification pairs from CFP-FP and calculate the cosine distance (\textit{angle}) between these pairs predicted by different models trained from the CASIA and combined data. Intuitively, the angles between these challenging pairs are significantly reduced when generated data are used for the model training.


\begin{table}[t]
\begin{center}
\begin{tabular}{c|c|c|c|c|c}
\hline
Training Data
&
{\includegraphics[width=0.05\textwidth]{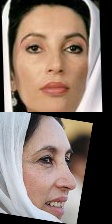}}
&
{\includegraphics[width=0.05\textwidth]{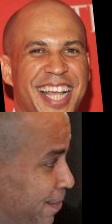}}
&
{\includegraphics[width=0.05\textwidth]{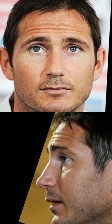}}
&
{\includegraphics[width=0.05\textwidth]{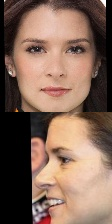}}
&
{\includegraphics[width=0.05\textwidth]{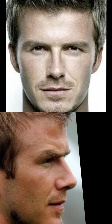}}\\
\hline
CASIA         & $84.06^{\circ}$ & $82.39^{\circ}$ & $84.72^{\circ}$ & $88.06^{\circ}$ & $84.37^{\circ}$  \\
CASIA+Gen    & $57.60^{\circ}$ & $63.12^{\circ}$ & $66.10^{\circ}$ & $59.72^{\circ}$ & $60.25^{\circ}$  \\
\hline
\end{tabular}
\end{center}
\caption{The angles between face pairs from CFP-FP predicted by different models trained from the CASIA and combined data. The generated data can obviously enhance the pose-invariant feature embedding.}
\label{tab:compactness}
\end{table}

\section{Conclusion}
We presented the first 3D face model for joint texture, shape, and normal generation based on Generative Adversarial Networks (GANs). The proposed GAN model implements a new architecture for exploiting the correlation between different modalities and can synthesize different facial expressions in accordance with the embeddings of an expression recognition network. We demonstrate that randomly synthesized images of our unified generator show strong relations between texture, shape, and normals and that rendering with normals provides excellent shading and overall visual quality. Finally, in order to demonstrate the generalization of our model, we have used a set of generated images to train a deep face recognition network. 

\section*{Acknowledgements}
Baris Gecer is supported by the Turkish Ministry of National Education, Stylianos
Ploumpis by the EPSRC Project EP/N007743/1 (FACER2VM), and Stefanos Zafeiriou by EPSRC Fellowship DEFORM (EP/S010203/1).

\clearpage
%
%
\bibliographystyle{splncs04}
\bibliography{egbib}

\begin{thebibliography}{10}
\providecommand{\url}[1]{\texttt{#1}}
\providecommand{\urlprefix}{URL }
\providecommand{\doi}[1]{https://doi.org/#1}

\bibitem{akimoto1993automatic}
Akimoto, T., Suenaga, Y., Wallace, R.S.: Automatic creation of {{3D}} facial
  models. IEEE Computer Graphics and Applications  \textbf{13}(5),  16--22
  (1993). \doi{10.1109/38.232096}

\bibitem{amberg2008expression}
Amberg, B., Knothe, R., Vetter, T.: Expression invariant {{3D}} face
  recognition with a morphable model. In: 2008 8th {{IEEE}} International
  Conference on Automatic Face and Gesture Recognition, {{FG}} 2008. pp.~1--6.
  {IEEE} (2008). \doi{10.1109/AFGR.2008.4813376}

\bibitem{amberg2007optimal}
Amberg, B., Romdhani, S., Vetter, T.: Optimal step nonrigid {{ICP}} algorithms
  for surface registration. In: Proceedings of the {{IEEE}} Computer Society
  Conference on Computer Vision and Pattern Recognition. pp.~1--8 (2007).
  \doi{10.1109/CVPR.2007.383165}

\bibitem{bao2018towards}
Bao, J., Chen, D., Wen, F., Li, H., Hua, G.: Towards open-set identity
  preserving face synthesis. In: Proceedings of the {{IEEE}} Computer Society
  Conference on Computer Vision and Pattern Recognition. pp. 6713--6722 (2018).
  \doi{10.1109/CVPR.2018.00702}

\bibitem{bazrafkan2018face}
Bazrafkan, S., Javidnia, H., Corcoran, P.: Face synthesis with landmark points
  from generative adversarial networks and inverse latent space mapping. arXiv
  preprint arXiv:1802.00390  (2018)

\bibitem{blanz1999morphable}
Blanz, V., Vetter, T.: A morphable model for the synthesis of {{3D}} faces. In:
  Proceedings of the 26th Annual Conference on Computer Graphics and
  Interactive Techniques, {{SIGGRAPH}} 1999. pp. 187--194. {ACM
  Press/Addison-Wesley Publishing Co.} (1999). \doi{10.1145/311535.311556}

\bibitem{booth2018large}
Booth, J., Roussos, A., Ponniah, A., Dunaway, D., Zafeiriou, S.: Large scale
  {{3D}} morphable models. International Journal of Computer Vision
  \textbf{126}(2-4),  233--254 (2018). \doi{10.1007/s11263-017-1009-7}

\bibitem{booth20163d}
Booth, J., Roussos, A., Zafeiriou, S., Ponniahy, A., Dunaway, D.: A {{3D}}
  morphable model learnt from 10,000 faces. In: Proceedings of the {{IEEE}}
  Computer Society Conference on Computer Vision and Pattern Recognition. vol.
  2016-December, pp. 5543--5552 (2016). \doi{10.1109/CVPR.2016.598}

\bibitem{bouaziz2013online}
Bouaziz, S., Wang, Y., Pauly, M.: Online modeling for realtime facial
  animation. ACM Transactions on Graphics  \textbf{32}(4), ~40 (2013).
  \doi{10.1145/2461912.2461976}

\bibitem{breidt2011robust}
Breidt, M., Biilthoff, H.H., Curio, C.: Robust semantic analysis by synthesis
  of {{3D}} facial motion. In: 2011 {{IEEE}} International Conference on
  Automatic Face and Gesture Recognition and Workshops, {{FG}} 2011. pp.
  713--719. {IEEE} (2011). \doi{10.1109/FG.2011.5771336}

\bibitem{brock2018large}
Brock, A., Donahue, J., Simonyan, K.: Large scale {{GaN}} training for high
  fidelity natural image synthesis. 7th International Conference on Learning
  Representations, ICLR 2019  (2019)

\bibitem{brunton2014review}
Brunton, A., Salazar, A., Bolkart, T., Wuhrer, S.: Review of statistical shape
  spaces for {{3D}} data with comparative analysis for human faces. Computer
  Vision and Image Understanding  \textbf{128},  1--17 (2014).
  \doi{10.1016/j.cviu.2014.05.005}

\bibitem{cao2013facewarehouse}
Cao, C., Weng, Y., Zhou, S., Tong, Y., Zhou, K.: {{FaceWarehouse}}: {{A 3D}}
  facial expression database for visual computing. IEEE Transactions on
  Visualization and Computer Graphics  \textbf{20}(3),  413--425 (2014).
  \doi{10.1109/TVCG.2013.249}

\bibitem{chen2019photo}
Chen, A., Chen, Z., Zhang, G., Mitchell, K., Yu, J.: Photo-realistic facial
  details synthesis from single image. In: Proceedings of the {{IEEE}}
  International Conference on Computer Vision. vol. 2019-October, pp.
  9428--9438 (Oct 2019). \doi{10.1109/ICCV.2019.00952}

\bibitem{cheng2019meshgan}
Cheng, S., Bronstein, M., Zhou, Y., Kotsia, I., Pantic, M., Zafeiriou, S.:
  {{MeshGAN}}: {{Non}}-linear {{3D}} morphable models of faces. arXiv preprint
  arXiv:1903.10384  (2019)

\bibitem{cootes2001active}
Cootes, T.F., Edwards, G.J., Taylor, C.J.: Active appearance models. Lecture
  Notes in Computer Science (including subseries Lecture Notes in Artificial
  Intelligence and Lecture Notes in Bioinformatics)  \textbf{1407}(6),
  484--498 (1998). \doi{10.1007/BFb0054760}

\bibitem{de2010optimal}
De~Smet, M., Van~Gool, L.: Optimal regions for linear model-based {{3D}} face
  reconstruction. In: Lecture Notes in Computer Science (Including Subseries
  Lecture Notes in Artificial Intelligence and Lecture Notes in
  Bioinformatics). vol. 6494 LNCS, pp. 276--289 (2011).
  \doi{10.1007/978-3-642-19318-7_22}

\bibitem{decarlo1998anthropometric}
DeCarlo, D., Metaxas, D., Stone, M.: An anthropometric face model using
  variational techniques. In: Proceedings of the 25th Annual Conference on
  Computer Graphics and Interactive Techniques, {{SIGGRAPH}} 1998. vol.~98, pp.
  67--74 (1998). \doi{10.1145/280814.280823}

\bibitem{deng2018uv}
Deng, J., Cheng, S., Xue, N., Zhou, Y., Zafeiriou, S.: {{UV}}-{{GAN}}:
  {{Adversarial}} facial {{UV}} map completion for pose-invariant face
  recognition. Proceedings of the IEEE Computer Society Conference on Computer
  Vision and Pattern Recognition pp. 7093--7102 (2018).
  \doi{10.1109/CVPR.2018.00741}

\bibitem{deng2018arcface}
Deng, J., Guo, J., Xue, N., Zafeiriou, S.: {{ArcFace}}: {{Additive}} angular
  margin loss for deep face recognition. Proceedings of the IEEE Computer
  Society Conference on Computer Vision and Pattern Recognition
  \textbf{2019-June},  4685--4694 (2019). \doi{10.1109/CVPR.2019.00482}

\bibitem{deng2017marginal}
Deng, J., Zhou, Y., Zafeiriou, S.: Marginal loss for deep face recognition. In:
  {{IEEE}} Computer Society Conference on Computer Vision and Pattern
  Recognition Workshops. vol. 2017-July, pp. 2006--2014 (2017).
  \doi{10.1109/CVPRW.2017.251}

\bibitem{gecer2017learning}
Gecer, B., Balntas, V., Kim, T.K.: Learning {{Deep Convolutional Embeddings}}
  for {{Face Representation Using Joint Sample}}- and {{Set}}-{{Based
  Supervision}}. In: 2017 {{IEEE International Conference}} on {{Computer
  Vision Workshops}} ({{ICCVW}}). pp. 1665--1672 (Oct 2017).
  \doi{10.1109/ICCVW.2017.195}

\bibitem{gecer2018facegan}
Gecer, B., Bhattarai, B., Kittler, J., Kim, T.K.: Semi-supervised {{Adversarial
  Learning}} to {{Generate Photorealistic Face Images}} of {{New Identities}}
  from {{3D Morphable Model}}. In: European {{Conference}} on {{Computer
  Vision}} ({{ECCV}}). vol. 11215, pp. 230--248. {Springer International
  Publishing} (2018). \doi{10.1007/978-3-030-01252-6_14}

\bibitem{gecer2019ganfit}
Gecer, B., Ploumpis, S., {Irene Kotsia}, Zafeiriou, S.: {{GANFIT}}:
  {{Generative Adversarial Network Fitting}} for {{High Fidelity 3D Face
  Reconstruction}}. In: Conference on {{Computer Vision}} and {{Pattern
  Recognition}} ({{CVPR}}). pp. 1155--1164 (Jun 2019).
  \doi{10.1109/CVPR.2019.00125}

\bibitem{goodfellow2014generative}
Goodfellow, I.J., {Pouget-Abadie}, J., Mirza, M., Xu, B., {Warde-Farley}, D.,
  Ozair, S., Courville, A., Bengio, Y.: Generative adversarial nets. In:
  Advances in Neural Information Processing Systems. vol.~3, pp. 2672--2680
  (2014). \doi{10.3156/jsoft.29.5_177_2}

\bibitem{gower1975generalized}
Gower, J.C.: Generalized procrustes analysis. Psychometrika  \textbf{40}(1),
  33--51 (1975). \doi{10.1007/BF02291478}

\bibitem{gulrajani2017improved}
Gulrajani, I., Ahmed, F., Arjovsky, M., Dumoulin, V., Courville, A.: Improved
  training of wasserstein {{GANs}}. In: Advances in Neural Information
  Processing Systems. vol. 2017-December, pp. 5768--5778 (2017)

\bibitem{he2016deep}
He, K., Zhang, X., Ren, S., Sun, J.: Deep residual learning for image
  recognition. In: Proceedings of the {{IEEE}} Computer Society Conference on
  Computer Vision and Pattern Recognition. vol. 2016-December, pp. 770--778
  (2016). \doi{10.1109/CVPR.2016.90}

\bibitem{hu2017avatar}
Hu, L., Saito, S., Wei, L., Nagano, K., Seo, J., Fursund, J., Sadeghi, I., Sun,
  C., Chen, Y.C., Li, H.: Avatar digitization from a single image for real-time
  rendering. ACM Transactions on Graphics  \textbf{36}(6), ~195 (2017).
  \doi{10.1145/3130800.3130887}

\bibitem{hu2018pose}
Hu, Y., Wu, X., Yu, B., He, R., Sun, Z.: Pose-guided photorealistic face
  rotation. In: Proceedings of the {{IEEE}} Computer Society Conference on
  Computer Vision and Pattern Recognition. pp. 8398--8406 (2018).
  \doi{10.1109/CVPR.2018.00876}

\bibitem{jamaludin2019you}
Jamaludin, A., Chung, J.S., Zisserman, A.: You said that?: {{Synthesising}}
  talking faces from audio. International Journal of Computer Vision
  \textbf{127}(11-12),  1767--1779 (2019). \doi{10.1007/s11263-019-01150-y}

\bibitem{karras2017progressive}
Karras, T., Aila, T., Laine, S., Lehtinen, J.: Progressive growing of {{GANs}}
  for improved quality, stability, and variation. In: 6th International
  Conference on Learning Representations, {{ICLR}} 2018 - Conference Track
  Proceedings (2018)

\bibitem{karras2018style}
Karras, T., Laine, S., Aila, T.: A style-based generator architecture for
  generative adversarial networks. Proceedings of the IEEE Computer Society
  Conference on Computer Vision and Pattern Recognition  \textbf{2019-June},
  4396--4405 (2019). \doi{10.1109/CVPR.2019.00453}

\bibitem{kingma2013auto}
Kingma, D.P., Welling, M.: Auto-encoding variational bayes. 2nd International
  Conference on Learning Representations, ICLR 2014 - Conference Track
  Proceedings  (2014)

\bibitem{lattas2020avatarme}
Lattas, A., Moschoglou, S., Gecer, B., Ploumpis, S., Triantafyllou, V., Ghosh,
  A., Zafeiriou, S.: {{AvatarMe}}: {{Realistically Renderable 3D Facial
  Reconstruction}} "in-the-wild". In: Conference on {{Computer Vision}} and
  {{Pattern Recognition}} ({{CVPR}}). pp. 760--769 (2020)

\bibitem{li2010example}
Li, H., Weise, T., Pauly, M.: Example-based facial rigging. ACM SIGGRAPH 2010
  Papers, SIGGRAPH 2010  \textbf{29}(4), ~32 (2010).
  \doi{10.1145/1778765.1778769}

\bibitem{li2012data}
Li, K., Dai, Q., Wang, R., Liu, Y., Xu, F., Wang, J.: A data-driven approach
  for facial expression retargeting in video. In: {{IEEE}} Transactions on
  Multimedia. vol.~16, pp. 299--310. {IEEE} (2014).
  \doi{10.1109/TMM.2013.2293064}

\bibitem{li2017learning}
Li, T., Bolkart, T., Black, M.J., Li, H., Romero, J.: Learning a model of
  facial shape and expression from {{4D}} scans. ACM Transactions on Graphics
  \textbf{36}(6), ~194 (2017). \doi{10.1145/3130800.3130813}

\bibitem{litany2018deformable}
Litany, O., Bronstein, A., Bronstein, M., Makadia, A.: Deformable shape
  completion with graph convolutional autoencoders. In: Proceedings of the
  {{IEEE}} Computer Society Conference on Computer Vision and Pattern
  Recognition. pp. 1886--1895 (2018). \doi{10.1109/CVPR.2018.00202}

\bibitem{Marmoset19:Toolbag}
{Marmoset LLC}: Marmoset toolbag (2019)

\bibitem{masi2016we}
Masi, I., Tr{\^a}n, A.T., Hassner, T., Leksut, J.T., Medioni, G.: Do we really
  need to collect millions of faces for effective face recognition? In: Lecture
  Notes in Computer Science (Including Subseries Lecture Notes in Artificial
  Intelligence and Lecture Notes in Bioinformatics). vol. 9909 LNCS, pp.
  579--596. {Springer} (2016). \doi{10.1007/978-3-319-46454-1_35}

\bibitem{mohammed2009visio}
Mohammed, U., Prince, S.J., Kautz, J.: Visio-lization: {{Generating}} novel
  facial images. ACM Transactions on Graphics  \textbf{28}(3), ~57 (2009).
  \doi{10.1145/1531326.1531363}

\bibitem{Moschoglou2017AgeDB}
Moschoglou, S., Papaioannou, A., Sagonas, C., Deng, J., Kotsia, I., Zafeiriou,
  S.: {{AgeDB}}: {{The}} first manually collected, in-the-wild age database.
  In: {{IEEE}} Computer Society Conference on Computer Vision and Pattern
  Recognition Workshops. vol. 2017-July, pp. 1997--2005 (2017).
  \doi{10.1109/CVPRW.2017.250}

\bibitem{moschoglou20193dfacegan}
Moschoglou, S., Ploumpis, S., Nicolaou, M., Papaioannou, A., Zafeiriou, S.:
  {{3DFaceGAN}}: {{Adversarial}} nets for {{3D}} face representation,
  generation, and translation. arXiv preprint arXiv:1905.00307  (2019)

\bibitem{odena2017conditional}
Odena, A., Olah, C., Shlens, J.: Conditional image synthesis with auxiliary
  classifier {{GANs}}. In: Proceedings of the 34th {{International Conference}}
  on {{Machine Learning}} - {{Volume}} 70. pp. 2642--2651. {{ICML}}'17,
  {JMLR.org}, {Sydney, NSW, Australia} (Aug 2017)

\bibitem{patel20093d}
Patel, A., Smith, W.A.: {{3D}} morphable face models revisited. In: 2009
  {{IEEE}} Computer Society Conference on Computer Vision and Pattern
  Recognition Workshops, {{CVPR}} Workshops 2009. vol. 2009 IEEE Computer
  Society Conference on Computer Vision and Pattern Recognition, pp.
  1327--1334. {IEEE} (2009). \doi{10.1109/CVPRW.2009.5206522}

\bibitem{paysan20093d}
Paysan, P., Knothe, R., Amberg, B., Romdhani, S., Vetter, T.: A {{3D Face
  Model}} for {{Pose}} and {{Illumination Invariant Face Recognition}}. In:
  2009 {{Sixth IEEE International Conference}} on {{Advanced Video}} and
  {{Signal Based Surveillance}}. pp. 296--301 (Sep 2009).
  \doi{10.1109/AVSS.2009.58}

\bibitem{platt1981animating}
Platt, S.M., Badler, N.I.: Animating facial expressions. In: Proceedings of the
  8th Annual Conference on Computer Graphics and Interactive Techniques,
  {{SIGGRAPH}} 1981. vol.~15, pp. 245--252. {ACM} (1981).
  \doi{10.1145/800224.806812}

\bibitem{ploumpis2020complete}
Ploumpis, S., Ververas, E., O'~Sullivan, E., Moschoglou, S., Wang, H., Pears,
  N., Smith, W.A.P., Gecer, B., Zafeiriou, S.: Towards a complete {{3D}}
  morphable model of the human head. IEEE Transactions on Pattern Analysis and
  Machine Intelligence (TPAMI) pp.~1--1 (2020).
  \doi{10.1109/TPAMI.2020.2991150}

\bibitem{ploumpis2019combining}
Ploumpis, S., Wang, H., Pears, N., Smith, W.A., Zafeiriou, S.: Combining {{3D}}
  morphable models: {{A}} large scale face-and-head model. In: Proceedings of
  the {{IEEE}} Computer Society Conference on Computer Vision and Pattern
  Recognition. vol. 2019-June, pp. 10926--10935 (2019).
  \doi{10.1109/CVPR.2019.01119}

\bibitem{pumarola2018ganimation}
Pumarola, A., Agudo, A., Martinez, A.M., Sanfeliu, A., {Moreno-Noguer}, F.:
  {{GANimation}}: {{Anatomically}}-aware facial animation from a single image.
  In: Lecture Notes in Computer Science (Including Subseries Lecture Notes in
  Artificial Intelligence and Lecture Notes in Bioinformatics). vol. 11214
  LNCS, pp. 835--851 (2018). \doi{10.1007/978-3-030-01249-6_50}

\bibitem{radford2015unsupervised}
Radford, A., Metz, L., Chintala, S.: Unsupervised representation learning with
  deep convolutional generative adversarial networks. 4th International
  Conference on Learning Representations, ICLR 2016 - Conference Track
  Proceedings  (2016)

\bibitem{ranjan2018generating}
Ranjan, A., Bolkart, T., Sanyal, S., Black, M.J.: Generating {{3D}} faces using
  convolutional mesh autoencoders. Lecture Notes in Computer Science (including
  subseries Lecture Notes in Artificial Intelligence and Lecture Notes in
  Bioinformatics)  \textbf{11207 LNCS},  725--741 (2018).
  \doi{10.1007/978-3-030-01219-9_43}

\bibitem{sela2017unrestricted}
Sela, M., Richardson, E., Kimmel, R.: Unrestricted facial geometry
  reconstruction using image-to-image translation. In: Proceedings of the
  {{IEEE}} International Conference on Computer Vision. vol. 2017-October, pp.
  1585--1594 (2017). \doi{10.1109/ICCV.2017.175}

\bibitem{sengupta2016frontal}
Sengupta, S., Chen, J.C., Castillo, C., Patel, V.M., Chellappa, R., Jacobs,
  D.W.: Frontal to profile face verification in the wild. In: 2016 {{IEEE}}
  Winter Conference on Applications of Computer Vision, {{WACV}} 2016 (2016).
  \doi{10.1109/WACV.2016.7477558}

\bibitem{shen2018faceid}
Shen, Y., Luo, P., Yan, J., Wang, X., Tang, X.: {{FaceID}}-{{GAN}}:
  {{Learning}} a symmetry three-player {{GAN}} for identity-preserving face
  synthesis. In: Proceedings of the {{IEEE}} Computer Society Conference on
  Computer Vision and Pattern Recognition. pp. 821--830 (2018).
  \doi{10.1109/CVPR.2018.00092}

\bibitem{slossberg2018high}
Slossberg, R., Shamai, G., Kimmel, R.: High quality facial surface and texture
  synthesis via generative adversarial networks. Lecture Notes in Computer
  Science (including subseries Lecture Notes in Artificial Intelligence and
  Lecture Notes in Bioinformatics)  \textbf{11131 LNCS},  498--513 (2019).
  \doi{10.1007/978-3-030-11015-4_36}

\bibitem{thies2015real}
Thies, J., Zollh{\"o}fer, M., Nie{\ss}ner, M., Valgaerts, L., Stamminger, M.,
  Theobalt, C.: Real-time expression transfer for facial reenactment. ACM
  Transactions on Graphics  \textbf{34}(6),  181--183 (2015).
  \doi{10.1145/2816795.2818056}

\bibitem{tran2018extreme}
Tran, A.T., Hassner, T., Masi, I., Paz, E., Nirkin, Y., Medioni, G.: Extreme
  {{3D Face Reconstruction}}: {{Seeing Through Occlusions}}. In: 2018
  {{IEEE}}/{{CVF Conference}} on {{Computer Vision}} and {{Pattern
  Recognition}}. pp. 3935--3944. {IEEE}, {Salt Lake City, UT} (Jun 2018).
  \doi{10.1109/CVPR.2018.00414}

\bibitem{tran2018representation}
Tran, L., Yin, X., Liu, X.: Representation learning by rotating your faces.
  IEEE Transactions on Pattern Analysis and Machine Intelligence
  \textbf{41}(12),  3007--3021 (2019). \doi{10.1109/TPAMI.2018.2868350}

\bibitem{trigueros2018generating}
Trigueros, D.S., Meng, L., Hartnett, M.: Generating photo-realistic training
  data to improve face recognition accuracy. arXiv preprint arXiv:1811.00112
  (2018)

\bibitem{yang2011expression}
Yang, F., Metaxas, D., Wang, J., Shechtman, E., Bourdev, L.: Expression flow
  for {{3D}}-{{Aware}} face component transfer. ACM Transactions on Graphics
  \textbf{30}(4),  1--10 (2011). \doi{10.1145/2010324.1964955}

\bibitem{yi2014learning}
Yi, D., Lei, Z., Liao, S., Li, S.Z.: Learning face representation from scratch.
  arXiv:1411.7923  (2014)

\bibitem{yin2017towards}
Yin, X., Yu, X., Sohn, K., Liu, X., Chandraker, M.: Towards large-pose face
  frontalization in the wild. In: Proceedings of the {{IEEE}} International
  Conference on Computer Vision. vol. 2017-October, pp. 4010--4019 (2017).
  \doi{10.1109/ICCV.2017.430}

\bibitem{zhang2005geometry}
Zhang, Q., Liu, Z., Guo, B., Shum, H.: Geometry-driven photorealistic facial
  expression synthesis. Proceedings of the 2003 ACM SIGGRAPH/Eurographics
  Symposium on Computer Animation, SCA 2003  \textbf{12}(1),  48--60 (2003)

\bibitem{zhao2017dual}
Zhao, J., Xiong, L., Jayashree, K., Li, J., Zhao, F., Wang, Z., Pranata, S.,
  Shen, S., Yan, S., Feng, J.: Dual-agent {{GANs}} for photorealistic and
  identity preserving profile face synthesis. In: Advances in Neural
  Information Processing Systems. vol. 2017-December, pp. 66--76 (2017)

\bibitem{zhu2016face}
Zhu, X., Lei, Z., Liu, X., Shi, H., Li, S.Z.: Face alignment across large
  poses: {{A 3D}} solution. In: Proceedings of the {{IEEE}} Computer Society
  Conference on Computer Vision and Pattern Recognition. vol. 2016-December,
  pp. 146--155 (2016). \doi{10.1109/CVPR.2016.23}

\end{thebibliography}
\end{document}